\documentclass[final,5p,times,twocolumn,nopreprintline]{elsarticle}
\biboptions{numbers,sort&compress}

\usepackage{amsmath}
\usepackage{amssymb}
\usepackage{booktabs}
\usepackage{dsfont}
\usepackage{graphicx}
\usepackage{tabularx}
\usepackage{placeins}
\usepackage{float}
\usepackage{flushend}
\usepackage{algorithm}
\usepackage{algpseudocode}
\usepackage[hidelinks]{hyperref}
\usepackage{url}
\usepackage{xcolor}
\newif\ifanonymous
\anonymousfalse 
\newcommand{\AnonURL}[1]{\ifanonymous[link removed for double-anonymized review]\else\url{#1}\fi}
\newcolumntype{Y}{>{\raggedright\arraybackslash}X}
\newcommand{\PaperTableEdge}{\hspace{4pt}}
\newcommand{\PaperTableSetup}{%
  \vspace{4pt}%
  \small
  \setlength{\tabcolsep}{2.2pt}%
  \renewcommand{\arraystretch}{1.08}%
}

\begin{document}

\begin{frontmatter}
\title{Hallucination Risk Evaluation Framework for Question-Answering in Video-Language Models}

\ifanonymous
\author{Anonymous authors}
\else
\author{Guoxiang Ren\corref{cor1}}
\ead{z5611125@ad.unsw.edu.au}
\cortext[cor1]{Corresponding author.}
\author{Rohitash Chandra}
\address{Transitional Artificial Intelligence Research Group,
School of Mathematics and Statistics,
UNSW Sydney, NSW 2052, Australia}
\fi

\begin{abstract}
Video-language models (VLMs) can assign high confidence to incorrect answers, limiting the deployment of selective video question answering (VideoQA) strategies. This problem can worsen when target videos differ from the development data used to train a reliability estimator. We propose a black-box target-domain evaluation framework that improves the ranking of fixed VLM answers without updating the answering models.  
We encode cross-model agreement and cross-view variation in a typed response graph and derive two scores: a histogram-based gradient-boosting score trained on development data and a logistic-regression score trained on a labelled target pilot grouped by video. The pilot compares each candidate with the original selected-option probability; a candidate is deployed only when video-level resampling shows a positive, stable improvement; otherwise, the original score is retained. On a held-out 963-question HERBench split, we find that the method reduces mean model-specific area under the risk-coverage curve (AURC) by 16.64\%, with a minimum gain of 11.39\%. On a separate 911-question Perception Test split, the mean AURC reduction is 18.87\%, and the target check retains the original score for InternVL3.5 VLM. Our framework achieves the largest mean reduction in error among seven baselines that do not retrain the answering models. It ranks the risk of existing answers rather than detecting hallucinations, and most of the ranking signal comes from other model families. 

These results show that target-domain checking can improve VLM answer ranking for selective VideoQA deployment without retraining the answering models.
\end{abstract}


\end{frontmatter}

\section{Introduction}
\label{sec:introduction}

Video-language models can describe events and answer questions about actions,
temporal order, state changes, and causes.  Recent systems such as Qwen3-VL
\cite{bai2025qwen3} and InternVL3.5 \cite{wang2025internvl3} show broad
video-understanding ability.  Benchmarks such as Video-MME
\cite{fu2025video} and Perception Test \cite{patraucean2023perception} measure
overall accuracy, but accuracy alone does not tell a user which individual
answers are safe to trust.

This distinction matters because a plausible answer can be unsupported by the
video.  We use \emph{video hallucination} for such a failure, including an
answer that contradicts visible content or the order of events.  Existing
benchmarks document both content hallucination \cite{wang2024videohallucer}
and temporal hallucination \cite{li2025vidhalluc}.  A model can give
the correct option with modest confidence on one question, yet give a wrong
option with greater confidence on another.  Reporting accuracy alone hides
this ordering problem.

Selective prediction offers a practical response: accept answers judged
reliable and send uncertain cases for review \cite{geifman2017selective}.
The same idea has been used in visual question answering
\cite{whitehead2022reliable}.  The answer itself need not change; only the
order in which answers are trusted changes.  We measure this ordering with
the area under the risk-coverage curve (AURC): at each coverage level, risk
is the error rate among the answers kept so far \cite{zhou2025novel}.  Lower
AURC means that errors appear later.

When only some answers can be reviewed, a useful reliability score places
safer answers first and defers riskier ones.  The accepted subset can therefore
be safer even though the chosen options never change.  The central question is
not only whether a model is accurate overall, but also whether its confidence
identifies which individual answers are most likely to be correct.

The largest option probability is the simplest reliability score.  We call it
the \emph{original score}: it is the probability assigned to the model's chosen
answer.  An \emph{option-probability list} contains the probabilities of all
answer options, whereas a \emph{candidate score} is produced by a learned
reliability model and is not assumed to be a calibrated probability.  The
original score is cheap to obtain, but it describes only one response to one
sampling of the video.
Controlled temporal views can reveal whether the same answer survives a small
change in frame positions, a denser frame sample, or reversed frame order.
Responses from different model families add another source of evidence:
disagreement can expose uncertainty that a single model does not report
\cite{hamidieh2026complementing}.  Yet stability and agreement are not proof of
visual grounding.  In our VHD diagnostic set, several model families can
confidently choose the same answer suggested by general knowledge even when
the video shows the opposite.  We call this a shared, prior-driven error.

Lightweight correctness models can combine richer evidence than a single
probability.
Peer-based selectors, for example, use the behavior of several models to
improve decisions about which answers to accept
\cite{dancette2023improving}.  The main difficulty is transfer: uncertainty
estimates often deteriorate when the new data differ from the old data
\cite{ovadia2019can}, and a confidence estimator that works on one dataset
may fail on another \cite{cattelan2024fix}.  Richer features alone do
not by themselves justify replacing the original score on a new target
dataset.

We ask a narrow deployment question: given answers that a video-language model
has already produced, when does a new dataset provide enough evidence to
replace the original score with a learned reliability score?  We seek a better
ordering of the existing answers, not a more accurate answering model or a way
to revise answers after observing extra evidence.  The selection rule must
retain the original score whenever the learned ordering lacks target support.

The method records the available evidence in a structured temporal response graph.
Each node stores the probabilities from one model and one video view.  The
relations record whether two responses differ because the sampling changed or
because the model family changed.  A small classifier, applied after answer
generation, turns these measurements into a correctness score for the
unchanged base answer.  This black-box design requires neither hidden model
states nor further training of the answering model.  It does require
additional runs to collect the response set, so we report both ranking gains
and inference cost.

A graph score trained on development datasets can improve the average result
yet harm a particular combination of model and dataset, which we call an evaluation
setting.  Our initial transfer rule used unlabeled feature shifts and
leave-one-dataset-out development results to detect this risk.  Although it
improved the pre-revision Video-MME and EgoSchema evaluations, the
3,200-question Video-MME-v2 development test exposed an accepted setting whose
AURC deteriorated.  Feature similarity did not establish that a
development-trained score would continue to rank target errors late.  This
failure motivated the revised target check.

We therefore replace the indirect transfer test with a labeled target pilot
that always keeps questions from the same source video together.  For each
evaluation setting, the pilot compares the original score with a
development-trained HGB score and a target-pilot logistic-regression score that
uses all four views.  The target-pilot score is evaluated on held-out pilot
groups, ensuring that no pilot video is scored by a model trained on that
video.  Bootstrap resampling is likewise performed at the video level.  A
learned score is enabled only when its conservative improvement estimate
exceeds a threshold fixed during development; otherwise, the original score
is retained exactly.  Selection thus depends on observed target performance,
not feature similarity alone.

Four public datasets and VHD provide evidence for method development and
training.  The
Video-MME-v2 result is used only to develop the target-check pilot rule.  We
then conduct two held-out tests under prespecified protocols, using the same
three video-language-model families.  HERBench Lite-v2 tests
long videos across twelve tasks that require several pieces of evidence.  A
deterministic 500-video subset of Perception Test
\cite{patraucean2023perception} supplies a larger, independently split
short-video target.  Before inference on each target, we record the data split,
model revisions, candidate-score definitions, fitting procedures, success
rule.  This
separates method development from the evidence used for the final claim.

This study on video hallucination builds benchmarks and detectors for unsupported content \cite{wang2024videohallucer,li2025vidhalluc,gautam2026videohedge} or changes the model to reduce it \cite{bae2025mash}.  It does not tell a deployed system how far to trust an answer it has already given, and learned selectors for selective answering are usually fitted and tested on a single distribution.  We therefore study how to rank fixed video answers by risk and how labeled videos from a new target should decide whether a learned ranking is used there. 

The paper makes four contributions:

\begin{enumerate}
  \item a structured temporal response graph that combines complete option
  probabilities from several model families and deterministic video views
  while leaving the base answer unchanged; on the held-out targets, three base-view answers from different families reduce AURC by 14.11\% and 18.29\%, four views of one model by under 2\%;
  \item a target check that keeps complete videos together, compares reliability
  scores trained on the development datasets and the target pilot, and returns to the original
  score exactly when neither learned score is supported, with a Bonferroni-adjusted video-bootstrap lower bound as the criterion;
  \item a study that separates development from final held-out testing, uses the
  failure observed when transferring without target labels to revise the method, and tests that revision
  on two public targets against learned and training-free baselines; and
  \item VHD, a diagnostic dataset developed in this study to measure cases in which
  several model families confidently agree on the same wrong answer under
  visible, conflicting, and concealed temporal evidence.
\end{enumerate}

\begin{figure*}[t]
\centering
\includegraphics[width=\textwidth]{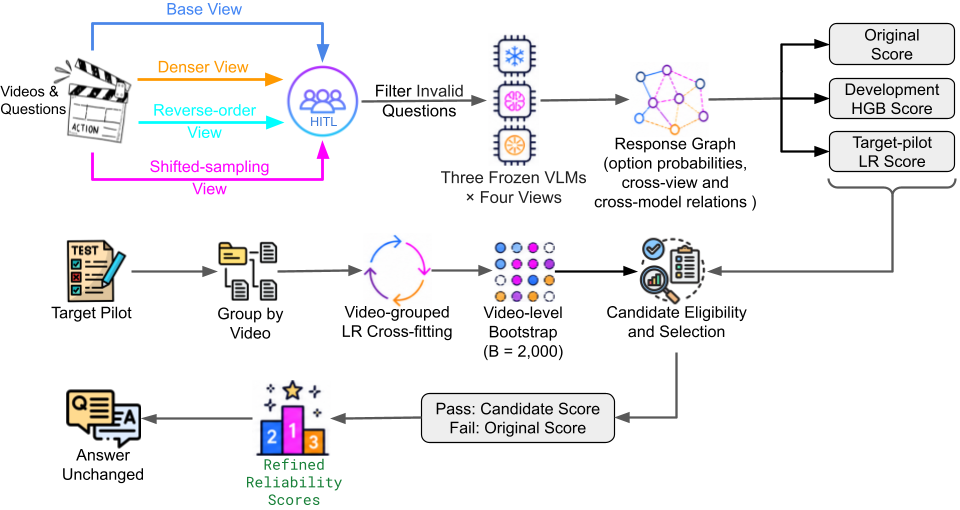}
\caption{Workflow for target-checked reliability score refinement.}
\label{fig:refinement-overview}
\end{figure*}
\section{Related work}
\label{sec:related}

\subsection{Reliability scores and selective answering}

Selective prediction ranks examples so that a system can defer uncertain
cases \cite{geifman2017selective}.  Coverage is the fraction of answers kept,
and risk is the error rate among those answers.  We use the area under the
risk-coverage curve (AURC) to summarize this trade-off.  AURC measures
ordering quality, not whether a numerical probability is perfectly calibrated
\cite{zhou2025novel}.  Temperature scaling is a standard way to adjust reported
confidence after training \cite{guo2017calibration}.  Such adjustments can
still deteriorate when the target data differ \cite{ovadia2019can}.  Work on
vision-language models likewise finds that calibration depends on the model,
data, and prompt \cite{tu2024empirical}.  A confidence estimator that improves
ranking without retraining the classifier must be checked again on
the new data \cite{cattelan2024fix}.

Work on reliable visual question answering (VQA) introduced abstention and
learned correctness selectors for image questions
\cite{whitehead2022reliable}.  Learning from Your Peers trains a
selector from several models and tests it on both familiar and shifted inputs
\cite{dancette2023improving}.  Adaptive Confidence Refinement uses learned
components to adjust the maximum softmax probability
\cite{tran2026knowing}.  These studies show that a learned selector can outperform
the original score.  We instead focus on transfer to a different dataset,
using observable option probabilities and requiring target evidence before
replacing the original score.

Other visual-question-answering methods learn richer uncertainty signals.
ViLU predicts failure from visual and language representations
\cite{lafon2025vilu}, while Variational VQA changes model training and uses
changes in its predicted distribution to decide which answers to defer
\cite{wieczorek2026variational}.  Our method works after the answering
models have been fixed.  It uses observable option probabilities, preserves the
base answer, and studies whether a learned reliability score should be used after
a move to a new dataset.

Other systems gather additional evidence while answering.  ReCoVERR asks
follow-up visual questions for low-confidence image answers
\cite{srinivasan2024selective}.  TRON samples sets of possible answers and
controls their error rate \cite{wang2025sample}.  VideoHEDGE instead
groups generations from clean and perturbed video clips
\cite{gautam2026videohedge}.  PhaseGuard+ samples one frozen model at several temporal offsets, answers by phase vote, and learns answer reliability \cite{liu2026phaseguard}; Section~\ref{sec:repeated-splits} compares it with our method, and in contrast, our method neither constructs
alternative-answer sets nor generates supporting text.  It ranks a fixed
multiple-choice answer using deterministic temporal views and a fixed set of
model families.

\subsection{Video hallucination and response consistency}

Video hallucination benchmarks test whether an answer is supported by the
visual input.  VideoHallucer separates errors about visible content from
unsupported additions
\cite{wang2024videohallucer}.  VidHalluc focuses more specifically on
temporal contradictions in video understanding \cite{li2025vidhalluc}.
Controlled evaluations also show that video-language-model answers can be inconsistent
across temporal-comprehension tests \cite{jung2025consistency}.  Conversely,
consistency itself is not proof of visual grounding: a stable or unanimous
answer can still ignore decisive evidence.

Consistency across similar visual questions can reveal black-box video-language model (VLM) failures
\cite{khan2024consistency}, and PARC measures structure across controlled
visual and language transformations \cite{schmalfuss2025parc}.  Cross-model
perplexity provides a training-free correctness signal without labels
\cite{gorbett2026cross}.  Semantic cross-model disagreement can also
complement self-consistency when one model is confidently wrong
\cite{hamidieh2026complementing}.  Our VHD experiment directly measures the case
in which a stable, unanimous answer remains wrong because the video contradicts
common expectations.

A supervised multi-model consensus engine has used answer-similarity graphs to
select a more likely answer from several language models
\cite{kallem2026learning}.  Our task is different: the method preserves
each video-language model's base answer and ranks that answer's risk from separately recorded
temporal and cross-model comparisons.

The response graph retains the operation that produced each response.  A phase
shift, a larger frame set, reversed order, and a different model family form
distinct relations.  This differs from graph calibration for communicating
agents, which studies how message exchange changes consensus
\cite{huang2026counterfactual}.  Our video-language models do not communicate; graph
features rank each model's unchanged answer rather than combine votes.

\subsection{Reliability scores on new datasets}

Selective question answering on a new dataset has used a separate
correctness model trained with labeled examples from the new domain
\cite{kamath2020selective}.  These results motivate checking whether the
original answer probability remains a useful score for deciding which answers
to accept.  We study
a related deployment decision: a labeled sample from the actual target data is
grouped by video and used to decide whether either of two fixed graph scores
should replace the original score.

Unlabeled target samples can support calibration when the input distribution
changes but the relationship between inputs and labels is assumed to remain
stable \cite{park2020calibrated}.  Feature similarity alone, however, cannot
show that a source ranking transfers.  Learn-then-test treats candidate
selection as several statistical tests and can give formal risk bounds for a
finite calibration set under its stated assumptions
\cite{angelopoulos2025learn}.  Related work extends these bounds to
covariate shift by weighting calibration examples \cite{almeida2025high}.

Recent question-answering systems also set acceptance thresholds using a
labeled set.  COIN controls how often accepted foundation-model answers are
wrong, including in multimodal settings \cite{wang2026coin}.  CIC controls
the error rate among accepted answers when calibration and test examples can
be treated as samples from the same distribution
\cite{dong2026uncertainty}.  These methods assume that an uncertainty score has already
been chosen.  We address an earlier decision: whether a learned ranking score
should replace the original score on a new dataset.

We test whether either predefined graph score improves AURC on a labeled
target pilot.  Complete videos remain grouped during cross-fitting and
bootstrap resampling, and candidate-specific minimum-improvement thresholds
determine whether a learned score replaces the original.  The resulting
interval describes the sampled target videos; unlike learn-then-test, it does
not provide a distribution-free guarantee.

 Table~\ref{tab:related-summary} compares the closest approaches to check whether they use  
responses from other models, temporal perturbations of the video, labelled data,
and whether they change the answer. 

\begin{table}[t]
\centering
\caption{Selected approaches compared by whether they use
responses from other models, temporal perturbations of the video, labeled
data, and whether they change the answer.}
\label{tab:related-summary}
\PaperTableSetup
\begin{tabularx}{\columnwidth}{@{\PaperTableEdge}Ycccc@{\PaperTableEdge}}
\toprule
Approach & \shortstack{Other\\models} & \shortstack{Temporal\\views} & Labels & \shortstack{Changes\\answer} \\
\midrule
Reliable VQA \cite{whitehead2022reliable} & no & no & yes & no \\
Learning from Your Peers \cite{dancette2023improving} & yes & no & yes & no \\
Selective QA under shift \cite{kamath2020selective} & no & no & yes & no \\
Cross-model perplexity \cite{gorbett2026cross} & yes & no & no & no \\
Semantic disagreement \cite{hamidieh2026complementing} & yes & no & no & no \\
VideoHEDGE \cite{gautam2026videohedge} & no & yes & no & no \\
PhaseGuard+ \cite{liu2026phaseguard} & no & yes & yes & yes \\
LLMs as a Jury \cite{liu2026jury} & yes & no & no & yes \\
This work & yes & yes & yes & no \\
\bottomrule
\end{tabularx}
\end{table}

\section{Methodology}
\label{sec:method}

\subsection{Problem setup and model outputs}

Let a model answer one multiple-choice question.  Under temporal
view \(v\), it assigns probability \(p_v(c)\) to option \(c\).  Here, \(v\)
identifies one of the four video views, \(c\) identifies an answer option, and
\(K\) is the number of options.  The complete probability list is
\[
  p_v=(p_v(1),\ldots,p_v(K)),\qquad
  p_v(c)\geq0,\qquad
  \sum_{c=1}^{K}p_v(c)=1.
\]
For the base view, written as \(v=0\), the answer returned to the user is
\[
  a=\arg\max_c p_0(c).
\]
The method leaves this answer unchanged and modifies only the score used to
rank or defer it.  Let \(b\) be the correct option.  The reliability model uses
the binary target
\[
  y=
  \begin{cases}
    1, & a=b,\\
    0, & a\ne b.
  \end{cases}
\]
For each question, we record a set of model outputs under
four views: a uniform base view, a view sampled at slightly shifted positions,
a denser view, and a reverse-order view.  The base view is
the reference.  The other three test sensitivity to a small sampling change,
additional visual evidence, and temporal order.  These roles are fixed, while
the number of frames follows the fixed protocol for each evaluation.
Every view is produced deterministically from the video and uses no answer
label.

\subsection{Response graph across models and video views}

The response graph \(G\) contains the recorded responses, the pairs being
compared, and the relation type for each pair.  Each node stores the full
probability list from one model under one video view.  An edge either links the
same model across two views or two model families under the same view.  Keeping
these relations distinct separates changes induced by temporal sampling from
cross-model disagreement.  The graph is a structured record of comparisons,
not a message-passing neural network.

For this model, \(a\) is the base-view answer defined above.  Basic features include
maximum probability, the top-two margin, candidate count, and normalized
entropy
\[
 H(p)=-\frac{\sum_c p(c)\log p(c)}{\log K}.
\]
The top-two margin is the difference between the two largest option
probabilities.  Entropy measures how spread out the probabilities are: it is
near zero when one option dominates and near one when the options receive
similar probabilities.
To measure the difference between two probability lists, we use normalized
Jensen-Shannon divergence
\[
 \operatorname{JS}(p,q)=
 \frac{\operatorname{KL}(p\|r)+\operatorname{KL}(q\|r)}{2\log 2},
 \qquad r=(p+q)/2.
\]
Here \(r\) is the average probability list, and \(\operatorname{KL}\) denotes
Kullback--Leibler divergence.  The normalized divergence is symmetric, bounded,
and zero for identical lists.
Comparisons between model families record answer agreement, support for \(a\)
from the other models, Jensen-Shannon divergence, vote fraction, variation in
support, and how concentrated the model choices are.  For each non-base view
\(v\), changes across views for the same model include
\[
  \operatorname{JS}(p_0,p_v),\quad
  H(p_v)-H(p_0),\quad
  p_v(a)-p_0(a),
\]
and an indicator that the most likely option changes.  Summaries across the fixed model set repeat
the vote, support, and entropy statistics under each view.  If a
development record lacks a view, it receives a presence indicator and
zero-filled relation values.
Support from another model is the probability it assigns to the base-view
answer; vote fraction is the share of models selecting that answer.  Support
variation measures dispersion in these probabilities, whereas choice
concentration measures how strongly the model choices cluster on one option.
Concatenating these measurements yields the numerical feature vector \(z\)
used by the reliability model and on the final targets, \(z\) has 53 entries (Appendix~\ref{app:features}).

\subsection{Checking and selecting a score on target data}
\label{sec:target-check-rule}

The initial transfer rule used target measurements without correct answers.
Video-MME-v2 exposed its weakness: similar feature distributions did not ensure
that a development-trained score would rank target errors well.  We therefore
use a labeled subset of target videos as the pilot.  The base answer is fixed
before pilot evaluation, so the labels affect only reliability scoring.

We split a target dataset by source video, keeping every question from one
video in the same partition.  The pilot contains half of the videos.  For
each model, we compare two graph scores with the model's original score:
\begin{enumerate}
  \item a development-trained HGB score, produced by a histogram-based
  gradient-boosting classifier trained on the development datasets; and
  \item a target-pilot logistic-regression score, trained on the labeled
  target pilot using all four views.
\end{enumerate}
The development-trained tree score is a histogram-based gradient-boosting
classifier.  It is trained on graph-feature vectors with equal total sample
weight for each development dataset,
learning rate 0.05, 200 boosting iterations, at most 15 leaf nodes per tree,
a minimum of 30 samples per leaf, and $\ell_2$ regularization of 2.0.  Its
output is the estimated probability that the chosen answer is correct.  The
target-pilot candidate is a regularized logistic-regression model with median
imputation, feature rescaling, equal total weight for correct and incorrect
answers, and fixed regularization strength \(C=0.1\).  For cross-fitting, the
pilot is divided into five video groups: four train the model and the fifth is
scored, with each group serving once as the held-out fold.  Thus, every pilot
video is scored by a model that was not trained on that video.  The model is
refitted on the complete pilot only after passing the target check.  The fixed
regularization strength was not tuned on either held-out test.  A
development-trained logistic-regression score is reported only as an
experimental baseline: when selected during Video-MME-v2 development, it had a
negative mean held-out gain and was therefore excluded from deployment
selection.

Let \(r\) denote the original score, namely the probability assigned to the
chosen answer, and let \(s\) denote whichever candidate
graph score is currently being checked.  On the pilot we measure the relative
reduction in AURC,
\begin{equation}
  g = 100\,
  \frac{\operatorname{AURC}(r)-\operatorname{AURC}(s)}
       {\operatorname{AURC}(r)}.
\end{equation}
Positive \(g\) indicates better ranking than the original score, and the
factor of 100 expresses the relative change as a percentage.  We resample
complete videos with replacement
$B=2{,}000$ times and recompute \(g\).  For candidate $c$, let
$g_{c,1},\ldots,g_{c,B}$ denote the bootstrap gains and define the adjusted
one-sided lower bound as
\begin{equation}
  L_c = Q_{\alpha/m}\!\left(g_{c,1},\ldots,g_{c,B}\right),
\end{equation}
where $Q_q$ is the empirical $q$-quantile, $m=2$ is the number of candidates,
and the family-wise significance level is $\alpha=0.05$.  Thus, each lower
bound is the 2.5th percentile of its bootstrap distribution.  The target-pilot
logistic-regression score is eligible when this bound is strictly positive.
The development HGB score uses the stricter threshold fixed during development:
its lower bound must exceed 0.5 percentage points.  If both candidates are
eligible, the method chooses the one with the larger lower bound and breaks an
exact tie by candidate name.

 Formally, for a target with candidate set \(\mathcal{C}=\{r,s_{\mathrm{HGB}},s_{\mathrm{LR}}\}\) and thresholds \(\tau_{\mathrm{HGB}}=0.5\) and \(\tau_{\mathrm{LR}}=0\), the deployed score is \(s^\star=\arg\max_{c\in\mathcal{C}\setminus\{r\}:\,L_c>\tau_c} L_c\), and \(s^\star=r\) when no candidate qualifies.  The bootstrap treats pilot videos as independent draws from the target and holds the fitted candidates fixed, so its bounds ignore the variability of refitting.  Resampling whole videos allows questions from one video to be dependent.  Under these assumptions, the Bonferroni split keeps the chance that either candidate passes by sampling error alone near 5\%. 

If neither candidate passes, the returned score equals the original score for
every question, which prevents an unsupported candidate from changing the
ranking.  Video-level resampling supports a decision for target videos
represented by the pilot; a later data change requires a new target check.

Algorithm~\ref{alg:target-check} summarizes the complete target check.  The final
test labels do not enter any step before the decision on which score to use has
been fixed.

\begin{algorithm}[t]
\caption{Target check for refining reliability scores}
\label{alg:target-check}
\small
\begin{algorithmic}[1]
\Require development-trained HGB model; labeled target pilot grouped by video
\Require response records; original scores $r$; $m=2$ candidates
\Require $B=2{,}000$ video-level bootstrap replicates; family-wise level $\alpha=0.05$
\State Extract one numerical feature vector $z$ for each base answer
\State Score the development HGB candidate on the pilot
\State Cross-fit the target-pilot logistic-regression candidate by video group
\For{each candidate $c$}
  \State Resample complete pilot videos $B$ times and compute
  $g_{c,1},\ldots,g_{c,B}$
  \State Set $L_c=Q_{\alpha/m}\!\left(g_{c,1},\ldots,g_{c,B}\right)$
\EndFor
\State Keep the target-pilot logistic-regression score if its $L_c>0$
\State Keep the development HGB score if its $L_c>0.5$ percentage points
\If{at least one candidate remains}
  \State Select the candidate with the largest $L_c$; break an exact tie by
  candidate name
  \State If the target-pilot candidate is selected, refit it on the full pilot
\Else
  \State Return the original score for every question
\EndIf
\State \Return the selected score for each fixed base answer
\end{algorithmic}
\end{algorithm}

\subsection{Computational cost and data separation}

Let \(N\) be the number of questions, \(M\) the number of model families,
\(V\) the number of temporal views, and \(K\) the largest number of answer
options.  Collecting these recorded responses requires \(NMV\) model calls.  Once
their option probabilities have been saved, constructing the feature vectors with
each model evaluated in turn costs
\(O\!\left(NK[M^2+M(V-1)]\right)\): each model is compared with the
other models under the same view and with its own non-base views.  The saved model outputs use
\(O(NKMV)\) numbers.  Logistic scoring is linear in the feature dimension;
HGB scoring is linear in the number of visited tree nodes.  In our setting
\(M=3\), \(V=4\), and \(K\leq8\) across all development and evaluation
datasets (\(K\leq5\) on the final targets), so graph construction and score calculation are
small compared with the twelve model calls per question.

The method never replaces \(a\), so its accuracy is identical to that
of the base video-language model on every subset.  A change in AURC or
AUROC comes only from reordering fixed answers.

If no candidate passes the pilot test, the returned scores are pointwise
identical to the original scores.  Under the same deterministic tie rule, both
methods then have the same ranking, risk-coverage curve, AURC, AUROC, ECE, and
Brier score.  We verify this identity before reporting results.

Pilot labels are used to fit and select a reliability model.  Final-test labels
are stored in a separate file and are unavailable to either operation.  All questions from
one video remain on the same side of this split.  The decision on which score to use is fixed before the final answer file
is opened.

The arithmetic cost of graph construction is small compared with the model
calls, but the full response set still requires several model evaluations per
question.  The method also assumes a finite set of answer
candidates.  Video-level resampling supports deployment on targets represented
by the pilot; a later data change requires a new target check.
\section{Experimental design}
\label{sec:experiments}

\subsection{Datasets and their roles}

Table~\ref{tab:datasets} records the role of each dataset.  The
development datasets are NExT-GQA \cite{xiao2024can}, STAR
\cite{wu2021star}, MVBench \cite{li2024mvbench}, and TempCompass
\cite{liu2024tempcompass}.  Together with VHD, they provide the training and
development examples for the graph scores.  Video-MME
\cite{fu2025video} and EgoSchema \cite{mangalam2023egoschema} are pre-revision
tests of the initial rule and its lower-cost variant.  Video-MME-v2
\cite{fu2026video} was a larger stress test planned before its labels were
opened.  Its result showed that unlabeled target features were not enough to
decide whether a score transferred.  We use that result only to develop
the target-check procedure.  HERBench is the first final test of the revised method
\cite{ben2026herbench}, and Perception Test is the second
\cite{patraucean2023perception}; both use prespecified protocols.

\begin{table*}[t]
\centering
\caption{Dataset roles and video-grouping units used in the study.}
\label{tab:datasets}
\PaperTableSetup
\begin{tabular}{@{\PaperTableEdge}lrrrrl@{\PaperTableEdge}}
\toprule
Dataset & Questions & Video groups & Answer options & Model families & Role \\
\midrule
NExT-GQA & 500 & 500 & 5 & 3 & Method development and training \\
STAR & 500 & 500 & 4 & 3 & Method development and training \\
MVBench & 1,800 & 1,723 & 2 to 4 & 3 & Method development and training \\
VHD (ours) & 801 & 341 & 2 & 3 & Method development, training, and diagnostic analysis \\
TempCompass & 410 & 410 & 2 & 3 & Method development and training \\
Video-MME & 1,089 & 363 & 4 & 3 & Pre-revision evaluation \\
EgoSchema & 500 & 500 & 5 & 3 & Pre-revision efficiency evaluation \\
Video-MME-v2 & 3,200 & 800 & 2 to 8 & 3 & Target-check development \\
HERBench Lite-v2 & 1,971 & 68 & 4 to 5 & 3 & Final held-out test 1 \\
Perception Test & 1,888 & 500 & 3 & 3 & Final held-out test 2 \\
\bottomrule
\end{tabular}
\end{table*}

VHD is used as a controlled diagnostic dataset; the two public targets provide
independent held-out tests.  VHD's 341 source clips comprise 245 licensed
real-video segments and 96 VideoPhy-2 clips, covering six physical-event
categories.  VHD contains 801 two-option questions with final labels checked
by human reviewers: 333 control questions with clearly visible
evidence, 178 questions where the video contradicts common expectations, and
290 questions with key visual evidence hidden.  We refer to these three conditions
as clearly visible evidence, video contradicts expectation, and key evidence
hidden.  The questions where the video contradicts common expectations reverse an
expected event so that an answer suggested by language or common sense
conflicts with the video.  Questions and transformations derived from one clip
stay in the same split or resampled group.
The released labels, group assignments, and permitted media resources are
available through the VHD Kaggle release%
\footnote{\AnonURL{https://www.kaggle.com/datasets/mlopssss/video-hallucination-diagnosis}}.
The software used to label and review VHD is available in the annotation-code
repository%
\footnote{\AnonURL{https://github.com/transitional-ai/hallucinationdataset}}.

Both reviewers independently assessed all 1,155 candidate questions.  Before
adjudication, they agreed on question validity in 97.7\% of cases
(\(\kappa=0.945\)).  For task-applicable labels on questions that both marked
valid, agreement was 89.2\% for the correct option, 89.4\% for the
language-prior and video-evidence labels, and 90.1\% for the uncertainty label
(\(\kappa=0.785\) to 0.796).  After adjudication, 354 candidates were rejected
and 801 entered the final question bank.  Appendix Table~\ref{tab:vhd-annotation-agreement} gives
the label-wise counts.

We call a response unanimous when all three model families choose the same
base option.  We report three quantities: the share of unanimous items, the
share that are both unanimous and wrong, and the error rate among unanimous
items.  These quantities
characterize the failure pattern and remain separate from HERBench method
selection and the decision on which score to use.

\subsection{Models and evaluation settings}

The development experiments use Qwen3-VL-4B-Instruct
\cite{bai2025qwen3}, InternVL3.5-4B-Instruct
\cite{wang2025internvl3}, and LLaVA-OneVision-2-8B-Instruct
\cite{an2026llava}, abbreviated as LLaVA-OV2 below.  The HERBench model set
uses Qwen3-VL-8B-Instruct, InternVL3.5-14B-Instruct, and
LLaVA-OneVision-2-8B-Instruct.  Their repository revisions were recorded before
inference.  Thus, the final tests change both the dataset and the model size for
Qwen and InternVL, while retaining the same LLaVA model size.

Development view definitions and frame budgets vary by dataset
(Appendix~\ref{app:development-views}).  HERBench and Perception Test use 16,
8, 24, and 8 frames for the base, shifted, denser, and reverse-order views.  The base view samples the video uniformly.  The
shifted-sampling view moves the sampling positions by half a step, the denser
view adds frames, and the reverse-order view feeds uniformly sampled frames from the end to the
beginning.  These fixed views test sensitivity to sampling, extra evidence,
and temporal order; benchmark-wide full-resource accuracy is outside the
present comparison.

Each call scores the ordered option labels rather than generating free-form
text.  A softmax converts the model's label scores into probabilities for the
answer choices.  Gold answers are kept out of the inference records.

All methods use the same fixed base answers and saved response records.  We
compare the target-checked score with the original score and five learned
reliability scores.  Their inputs provide progressively richer evidence:
confidence features alone, cross-model relations in the base view, or the full
four-view response graph.  The confidence-only features are maximum
probability, top-two margin, probability spread, and number of answer choices;
the base-view score adds same-view cross-model comparisons.  Neither uses
alternate temporal views.  Temperature or Platt scaling would leave AURC unchanged, since both are monotone in the original score; the confidence-only score combines several features and can reorder answers.  The three target-pilot scores (confidence only,
base view, and all four views) use the same video-grouped folds and logistic-
regression settings.  Scores trained on the development datasets are applied
to the target data without refitting.
Reporting these scores without the target check also separates ranking quality
from the rule that decides which score to use.
To keep the information available to all three model families identical, every
comparison uses the same saved option probabilities.  Methods that require hidden
representations or retraining the answering model are discussed in
Section~\ref{sec:related}, but fall outside this protocol based only on saved outputs.

\subsection{HERBench protocol}
\label{sec:herbench-protocol}

HERBench Lite-v2 contains 1,971 multiple-choice questions from 68 long videos
and covers twelve tasks that require evidence from more than one moment
\cite{ben2026herbench}.  A deterministic split created without using labels assigns 34 complete videos
(1,008 questions) to the target pilot and the other 34 videos (963
questions) to the final test.  The split is balanced by the four source-video
collections.  Test answers are stored separately and are not supplied to the
pilot.

The fixed model set is Qwen3-VL-8B-Instruct,
InternVL3.5-14B-Instruct, and LLaVA-OneVision-2-8B-Instruct.  Repository
commits were recorded before inference.  Each question is evaluated
under four fixed views: a 16-frame base view, an 8-frame view sampled at slightly
shifted positions, a 24-frame denser view, and an 8-frame view presented in
reverse order.  This gives 23,652 records, one for each model, question, and
view.  The experiment evaluates reliability in this
controlled low-frame visual setting rather than full-resource benchmark
accuracy.

All model outputs are produced before either analysis stage begins.  Only pilot
labels are available during fitting and score selection, after which the
decision is stored with a digital checksum.  Before reading the test labels,
the final stage verifies that neither this record nor the reserved test-answer
file has changed.  The primary result is the mean relative AURC reduction
across the three models.  To
pass the final check, the mean gain must be at least 1\%, the 95\% lower
video-level bootstrap bound must be positive, and no model may fall
below $-1\%$.  At least one evaluation setting must use a learned score,
and every rejected setting must match the original score exactly.  AUROC, ECE,
Brier score, risk at 50\% coverage, and task-level AURC are secondary results.

Before any HERBench prediction was generated, the 50\% pilot and the rule for
choosing between the two candidate scores were fixed on Video-MME-v2
(Section~\ref{sec:target-check-development}).
\subsection{Perception Test protocol}
\label{sec:perception-protocol}

We next use the validation split of the Google DeepMind Perception Test
\cite{patraucean2023perception}.  Its multiple-choice annotations cover 5,260
validation videos.  Before running any model, we applied a fixed
deterministic ordering computed from the video identifiers and retained
the first 500 videos.  This selection used
neither question text, answer options, nor labels.  A second deterministic
ordering, created with a different fixed key,
assigned 250 complete videos to the labeled target sample and 250 to final
testing.  The
resulting subsets contain 977 and 911 questions, respectively.  The questions
span abstraction, memory, physics, and semantics.  For videos whose revealing
final frames are withheld by the benchmark, we stop at the official cutoff
frame.

The model revisions, four views, candidate-score definitions, fitting
procedures, minimum-improvement thresholds, and final check are unchanged from
HERBench, and the split seeds, analysis code, and success rule were recorded
before inference.  The experiment provides an independent test of the method on
a different short-video dataset.

\subsection{Evaluation and additional checks}

AURC first sorts answers by reliability.  At each coverage level it keeps that fraction of the highest
scores and measures the error rate among them; the average of these risks is
AURC, so lower values are better.
If \(e_j\) is 1 when the \(j\)-th answer in this order is wrong and 0
otherwise, the discrete calculation is
\[
  \operatorname{AURC}
  =\frac{1}{N}\sum_{j=1}^{N}
    \frac{\sum_{\ell=1}^{j}e_\ell}{j}.
\]
Hallucination risk refers to the risk that a fixed answer is not supported by the video, measured here by whether it is wrong.  Benchmark labels do not say why an answer is wrong, so this also covers other errors; VHD isolates one cause, priors that the video contradicts. 
For each held-out test, the main outcome is the relative AUROC reduction from
the original score, averaged equally over the three models.  We
do not combine effects across the two benchmarks.  We also report the area
under the receiver operating characteristic curve (AUROC), which is higher when
correct answers are ranked above incorrect ones.  The 15-bin expected
calibration error (ECE) measures whether reported confidence matches actual
correctness, using the average gap between confidence and observed accuracy.
The Brier score is the mean squared probability
error.  We further report risk at 50\% coverage, model-level results, and
task-level AURC\@.  We estimate final uncertainty with 10,000 paired resamples
of complete test videos: 34 for HERBench and 250 for Perception Test.  These
intervals describe the sampled videos; they do not treat the three model
families as a random sample of future video-language models.

After the final-test labels became available, we ran four secondary analyses
without changing either score-selection decision.  First, paired video
resampling compares the selected scores with each reported baseline; these
descriptive intervals play no role in method selection.  Second, we reduce the
pilot to 10, 20, 30, or 40\% of all target videos while keeping the original
test set fixed.  For each fraction, 20 deterministic video subsets are
evaluated for all three model settings; the original 50\% pilot is run once.  A
harmful use occurs when the pilot accepts a learned score whose final-test AURC
is worse than the original score.  Third, saved option probabilities are used
to refit target scores with smaller model and view sets, with both model-call
and frame counts reported.  Fourth, seven training-free scores represent
common confidence and consistency signals: top-two margin, negative entropy,
base-view cross-model vote and support, own-view stability, and vote or support
across all views.  Discrete vote scores use the original score only to break
exact ties.  None of these scores requires labels, fitting, or additional model
calls.  The secondary analyses assess robustness and efficiency without
redefining the prespecified target-check policy.

\section{Results}
\label{sec:results}

\subsection{Development results and the need for target checking}

The leave-one-dataset-out evaluation, in which each complete development dataset is
held out in turn, shows why we need a rule for deciding when to use a
learned reliability score.  The development-trained logistic-regression score used without a target check improves
mean AURC by 11.61\% but loses 16.48\% in its worst evaluation setting.
The pre-revision rule is more cautious: it uses the learned score in seven of
15 settings and keeps the original score in eight, giving a 6.95\% mean gain
with no negative setting.  Comparisons between model families provide
most of the signal.  Replacing both other models with copies of the model being evaluated
reduces the gain to 3.73\%, while shuffling the labels that identify temporal
relations reduces this graph score's gain from 11.61\% to 10.03\%.

\begin{table}[t]
\centering
\caption{Leave-one-dataset-out results on the five development datasets.}
\label{tab:development-ablation}
\PaperTableSetup
\begin{tabularx}{\columnwidth}{@{\PaperTableEdge}Yrrrr@{\PaperTableEdge}}
\toprule
Method & Mean gain & Worst & Positive & Negative \\
\midrule
Logistic score, all graph features & 11.61\% & $-16.48$\% & 12 & 3 \\
Permuted relation labels & 10.03\% & $-15.90$\% & 12 & 3 \\
Other models, base view & 9.84\% & $-13.19$\% & 12 & 3 \\
\textbf{Pre-revision rule} & \textbf{6.95\%} & \textbf{0.00\%} & \textbf{7} & \textbf{0} \\
Copied other-model responses & 3.73\% & $-7.98$\% & 10 & 5 \\
Original score & 0.00\% & 0.00\% & 0 & 0 \\
\bottomrule
\end{tabularx}
\end{table}

On the 178 questions where the video contradicts common expectations, 65.2\%
receive the same answer from all three model families and 55.6\% are unanimous
and wrong.  Among unanimous responses, the error rate is 85.3\% (95\%
video-level bootstrap interval, 78.6 to 91.5\%), compared with 31.2\% for
control questions with clearly visible
evidence and 16.5\% for questions with key visual evidence hidden.  Model accuracy on
video-conflict questions ranges
from 16.3\% to 37.6\%, while mean confidence on wrong answers ranges from
0.730 to 0.918.  Increasing the frame count from eight to sixteen changes
accuracy by only \(-0.6\) to \(+1.1\) percentage points.  In this controlled
case, agreement and additional frames alone do not reliably identify correct
answers.

\begin{figure*}[t]
\centering
\includegraphics[width=0.92\textwidth]{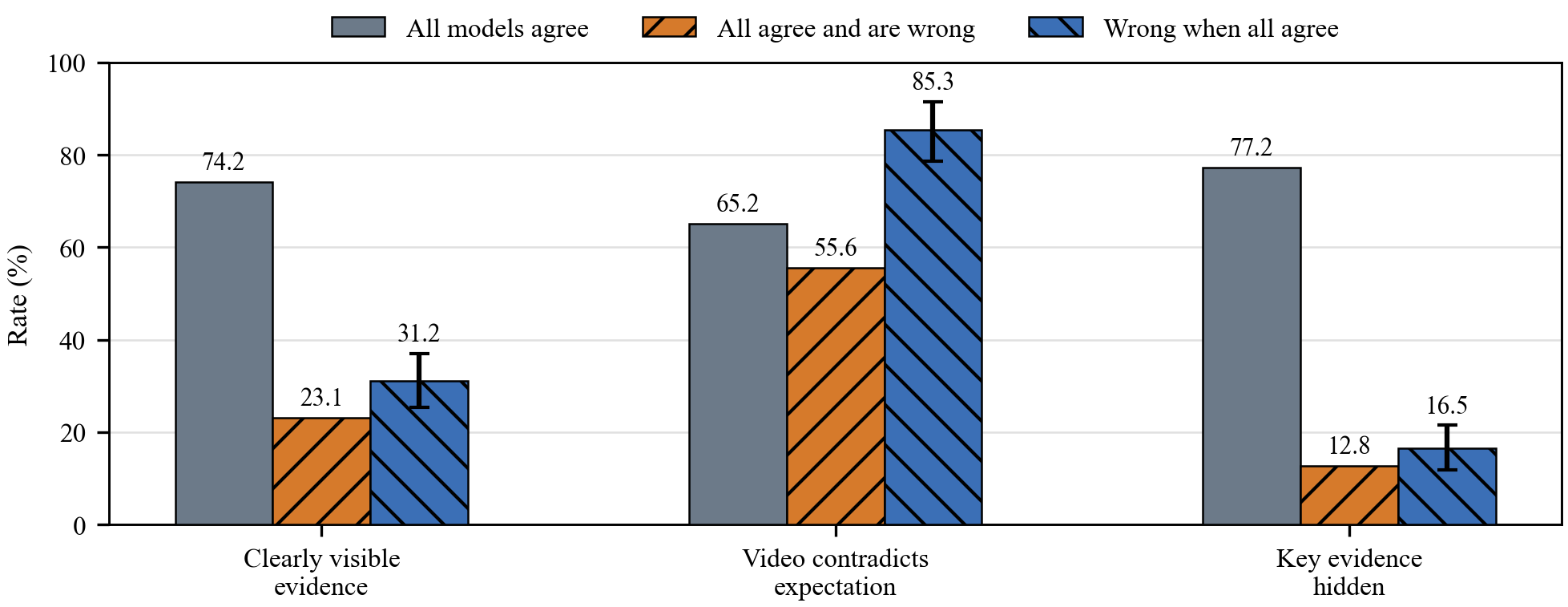}
\caption{VHD consensus diagnostic with 95\% source-clip bootstrap intervals.}
\label{fig:vhd-consensus}
\end{figure*}

Figure~\ref{fig:vhd-consensus} shows that unanimity itself is common in all
three conditions, but only the condition where the video contradicts common expectations combines it with a
high conditional error rate.

 We find that Logistic and HGB scores trained on the other four development datasets worsen AURC on the counterintuitive questions by 7.20\% and 5.48\%, with all of VHD held out.  For the 99 questions that all three models answer wrongly, they give mean confidence percentiles of 66.9 and 62.0, against 56.4 for the original score, so without target labels they trust these shared errors more (Appendix~\ref{app:vhd-holdout}).

The pre-revision tests used protocols specified before their labels became
available.  The pre-revision rule gained 10.95\% on Video-MME, while the
standard and lower-cost variants gained 19.81\% and 19.52\% on EgoSchema.  The
larger Video-MME-v2 run did not reproduce that pattern.  The pre-revision rule
gained only 0.09\% on average and lost 1.22\% in its worst model and dataset
setting; the lower-cost variant showed the same issue.
Table~\ref{tab:historical-panels} reports this failed row separately.

\begin{table}[t]
\centering
\caption{Results under protocols fixed before the target-check revision.}
\label{tab:historical-panels}
\PaperTableSetup
\begin{tabularx}{\columnwidth}{@{\PaperTableEdge}YYrrrr@{\PaperTableEdge}}
\toprule
Evaluation & Method & Settings & Mean & Worst & Learned \\
\midrule
Development & Pre-revision & 15 & 6.95\% & 0.00\% & 7 \\
Video-MME & Pre-revision & 3 & 10.95\% & 0.00\% & 2 \\
EgoSchema & Pre-revision & 3 & 19.81\% & 0.00\% & 2 \\
EgoSchema & Lower-cost & 3 & 19.52\% & 0.00\% & 2 \\
Video-MME-v2 & Pre-revision & 3 & 0.09\% & $-1.22$\% & 2 \\
Video-MME-v2 & Lower-cost & 3 & 0.10\% & $-1.34$\% & 2 \\
\bottomrule
\end{tabularx}
\end{table}

\subsection{Developing the target check on Video-MME-v2}
\label{sec:target-check-development}

We compared four pilot rules on 20 deterministic video-level splits.  Each row
in Table~\ref{tab:target-check-development} summarizes 60 evaluations across
splits and models at the selected 50\% pilot fraction.  The target-only rule
missed the required 0.5\% average gain.  Adding both development-trained
candidates passed the overall check but retained small losses in individual
cases.  Removing the development-trained logistic-regression candidate raised
both the mean gain and the fraction of positive cases, yet a permissive HGB
decision still produced a $-1.49\%$ loss.  The final rule therefore requires
the development HGB candidate's conservative lower estimate to exceed 0.5
percentage points, while retaining a zero threshold for the target-pilot
candidate.  This choice removed all losses on the development splits.

\begin{table}[t]
\centering
\caption{Video-MME-v2 target-check development at the 50\% pilot fraction.}
\label{tab:target-check-development}
\PaperTableSetup
\begin{tabularx}{\columnwidth}{@{\PaperTableEdge}Yrrrrl@{\PaperTableEdge}}
\toprule
Policy & Mean & Worst & Gain $>0$ & Original kept & Check \\
\midrule
Target pilot only & 0.47\% & $-0.62$\% & 30.0\% & 66.7\% & Fail \\
Three candidates & 0.92\% & $-0.65$\% & 28.3\% & 65.0\% & Pass \\
Two candidates, no margin & 0.96\% & $-1.49$\% & 33.3\% & 65.0\% & Fail \\
Final rule & 0.99\% & 0.00\% & 33.3\% & 66.7\% & Pass \\
\bottomrule
\end{tabularx}
\end{table}

The selected policy, split fraction, candidate margins, and HERBench success
rule were recorded before any HERBench prediction was generated.

The fraction was not chosen on HERBench.  Table~\ref{tab:pilot-size}
shows the three fractions examined during Video-MME-v2 development.  The
25\% and 35\% pilots caused no held-out losses but selected the graph score in too few cases:
each missed both the required 0.5\% average gain and 25\% positive rate.
The 50\% pilot was the only tested option that met the full development check.
The resulting protocol uses a labeled pilot of several hundred questions
rather than a few-shot sample.

\begin{table}[t]
\centering
\caption{Pilot-fraction sensitivity during Video-MME-v2 development.}
\label{tab:pilot-size}
\PaperTableSetup
\begin{tabular}{@{\PaperTableEdge}lrrrr@{\PaperTableEdge}}
\toprule
Pilot & Mean & Worst & Gain $>0$ & Original kept \\
\midrule
25\% & 0.32\% & 0.00\% & 13.3\% & 86.7\% \\
35\% & 0.33\% & 0.00\% & 13.3\% & 86.7\% \\
50\% & 0.99\% & 0.00\% & 33.3\% & 66.7\% \\
\bottomrule
\end{tabular}
\end{table}

\subsection{Held-out HERBench results}
\label{sec:herbench-results}

On the held-out HERBench split, the method meets the prespecified success
criteria.  Average AURC across the three models falls by 16.64\% (95\%
confidence interval (CI) from complete-video resampling, 12.11 to 22.59\%), and
the smallest model-level gain is 11.39\%.

\begin{table}[t]
\centering
\caption{HERBench final-test results (963 questions).}
\label{tab:herbench-main}
\PaperTableSetup
\begin{tabularx}{\columnwidth}{@{\PaperTableEdge}Yrrrrl@{\PaperTableEdge}}
\toprule
Model & Acc. & \shortstack[r]{Original\\AURC} & \shortstack[r]{Refined\\AURC} & Gain & \shortstack[l]{Selected\\score} \\
\midrule
InternVL3.5 & 0.464 & 0.3277 & 0.2904 & 11.39\% & target pilot \\
Qwen3-VL & 0.476 & 0.3727 & 0.2786 & 25.26\% & target pilot \\
LLaVA-OV2 & 0.481 & 0.3153 & 0.2734 & 13.29\% & target pilot \\
Model mean & 0.474 & 0.3385 & 0.2808 & 16.64\% & -- \\
\bottomrule
\end{tabularx}
\end{table}

\begin{figure*}[t]
\centering
\includegraphics[width=0.92\textwidth]{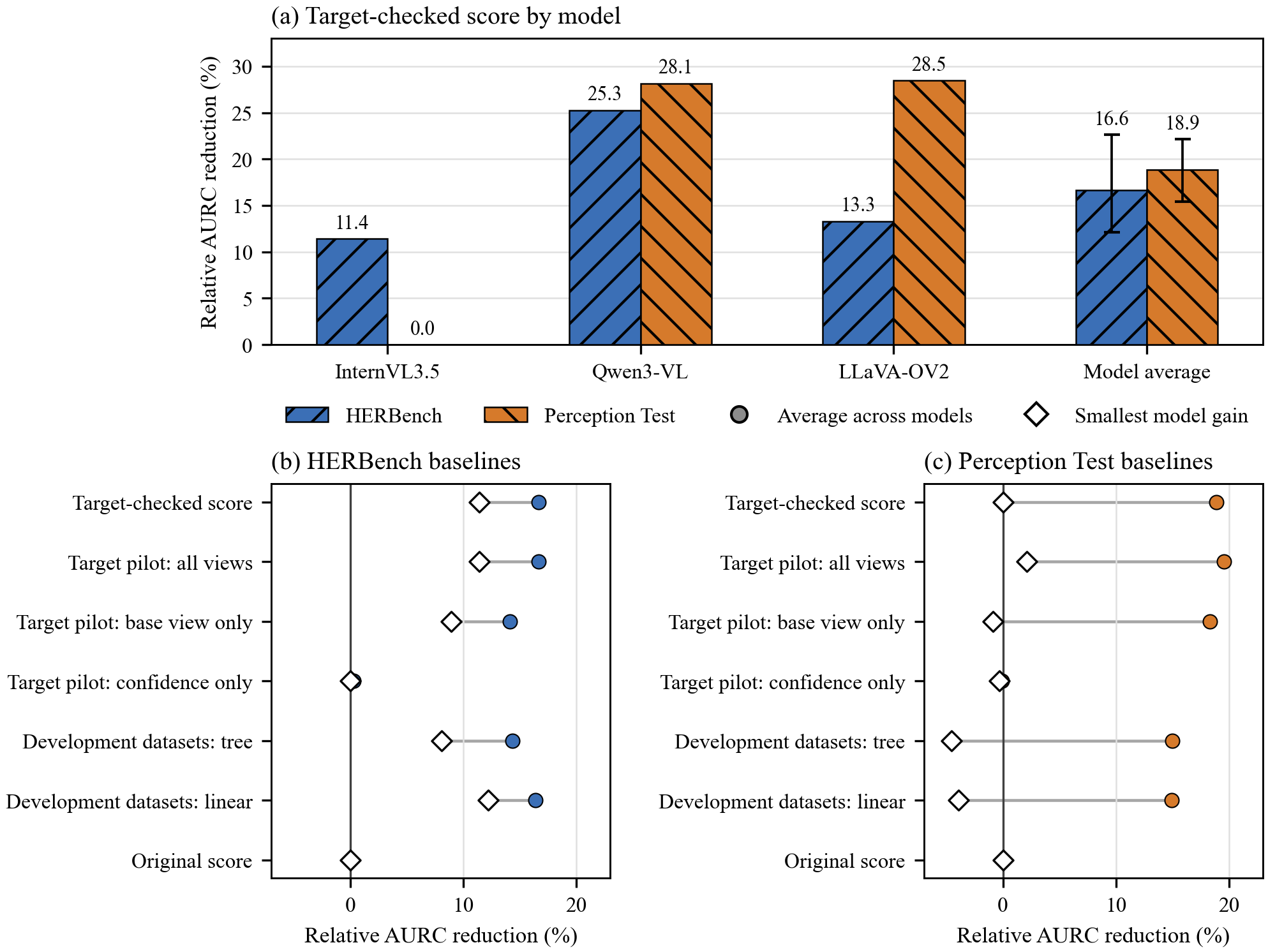}
\caption{Held-out results by model and scoring method.}
\label{fig:heldout-overview}
\end{figure*}

All three pilot decisions select the score trained on the target pilot using all four views
(Appendix~\ref{app:pilot-decisions}).  On the final split,
its mean gain is 2.54 percentage points larger than the score trained on the target pilot
using only the base view and 16.37 points larger than the score trained on the same pilot using confidence only
(Figure~\ref{fig:heldout-overview}(b)).  The development HGB score
is also strong on HERBench, but the earlier Video-MME-v2 loss shows why
evidence from the new target is needed before using it.  Results vary by task:
ten of twelve tasks have positive mean gains, while Scene Verification \&
Arrangement has a $-18.60\%$ mean change.  The method makes one decision for
each evaluation setting, not a separate decision for each task, so individual
tasks can still vary.  Both tasks with losses have accuracy below 0.40, and Scene Verification is the largest task (492 scored rows).  Accuracy alone does not explain the losses: Temporal Ordering, at 0.186, still gains 8.93\% (Appendix~\ref{app:herbench-tasks}).  A deployment dominated by such tasks would need its own pilot check.

\subsection{Held-out Perception Test results}
\label{sec:perception-results}

On the held-out Perception Test split, the method meets the prespecified
success criteria.  Average AURC across the three models falls by 18.87\% (95\%
CI from complete-video resampling, 15.41 to 22.12\%).  InternVL retains its
original score, so its AURC is unchanged.

\begin{table}[t]
\centering
\caption{Perception Test final-test results (911 questions).}
\label{tab:perception-main}
\PaperTableSetup
\begin{tabularx}{\columnwidth}{@{\PaperTableEdge}Yrrrrl@{\PaperTableEdge}}
\toprule
Model & Acc. & \shortstack[r]{Original\\AURC} & \shortstack[r]{Refined\\AURC} & Gain & \shortstack[l]{Selected\\score} \\
\midrule
InternVL3.5 & 0.728 & 0.1008 & 0.1008 & 0.00\% & original \\
Qwen3-VL & 0.673 & 0.1557 & 0.1119 & 28.14\% & target pilot \\
LLaVA-OV2 & 0.616 & 0.1895 & 0.1356 & 28.46\% & target pilot \\
Model mean & 0.672 & 0.1487 & 0.1161 & 18.87\% & -- \\
\bottomrule
\end{tabularx}
\end{table}

The conservative lower estimates
for Qwen and LLaVA exceed the required threshold (Appendix~\ref{app:pilot-decisions}), so both use the score trained
on the target pilot using all four views.  Their final AURC reductions are 28.14\% and 28.46\%, respectively.
InternVL does not meet the required threshold, so it retains its original
score and its AURC remains unchanged.  The four task groups
have gains, averaged across the three models, between 15.72\% and 22.04\%.

The score trained on
target-pilot confidence alone changes average AURC by
$-0.09\%$, whereas the score using only the base view gains 18.29\%
(Figure~\ref{fig:heldout-overview}(c)).  The difference shows
that relations between model responses provide information beyond ordinary
confidence features.
\subsection{Summary of held-out tests and additional checks}
\label{sec:two-confirmations}

Secondary metrics move in the same direction.  AUROC averaged across models
rises from 0.716 to 0.799 on HERBench and from 0.767 to 0.844 on Perception
Test, and ECE and Brier score fall on both targets
(Tables~\ref{tab:herbench-baselines} and~\ref{tab:perception-baselines}).
Against an oracle ordering, the target-checked score closes 33.4\% and 32.7\% of the original score's gap (Table~\ref{tab:eaurc}).
\label{sec:posthoc-robustness}

After the two held-out evaluations, we used the saved model outputs to examine
two practical questions: how pilot size affects the decision to enable a score
and how much of the response set is needed.  These checks keep the primary
scores, test splits, and decisions unchanged.

Paired video resampling compares the method with each baseline
(Table~\ref{tab:paired-baselines}).  The method improves on the development HGB
score on both targets, by 2.32 and 3.89 percentage points with intervals above
zero.  Temporal relations add 2.54 points beyond the base-view pilot score on
HERBench, with an interval above zero, but only 0.58 points on Perception Test.

On the reference test splits, the method also outperforms all seven
training-free baselines in AURC averaged across models.  The strongest such baseline on HERBench averages model
support over all views and reduces AURC by 15.57\%, close to the method's
16.64\%; the
paired difference is 1.08 percentage points (95\% CI, $-0.70$ to 2.91).  On
Perception Test, the strongest training-free score is base-view cross-model
support at 11.67\%, while the method reaches 18.87\%; the paired difference is
7.20 points (2.76 to 11.86).  Cross-model vote scores help on both datasets.  By
contrast, margin, entropy, and single-model temporal stability do not improve
the ranking on average.
Training-free support needs no labels, but without labels one cannot tell in advance where it is safe; on Perception Test, a 10\% pilot already beats every training-free score (Table~\ref{tab:pilot-sensitivity}).  Appendix Table~\ref{tab:training-free} reports
all scores and paired intervals.

\begin{figure*}[t]
\centering
\includegraphics[width=0.96\textwidth]{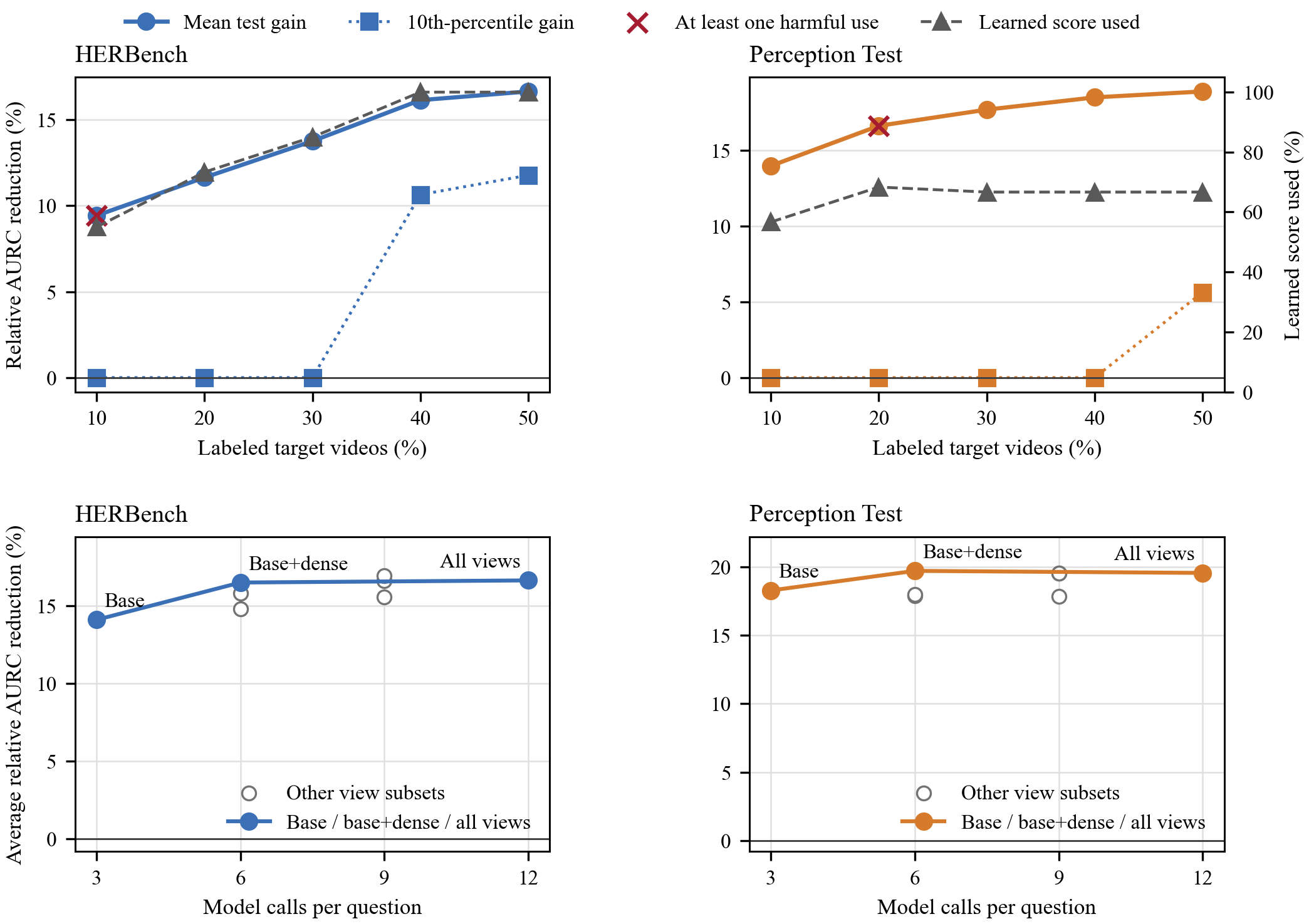}
\caption{Pilot-size sensitivity and model-input ablations on the fixed held-out
test splits.}
\label{fig:additional-efficiency}
\end{figure*}

The upper chart in Figure~\ref{fig:additional-efficiency} subsamples the
original pilot while keeping the test split fixed.  With 30\% or 40\% of target
videos labeled, mean test gains stay between 13.76\% and 18.48\%, and none of
the 60 evaluations per target at either fraction enables a harmful score.
Smaller pilots occasionally do: one of 60 HERBench evaluations at 10\% loses
5.84\%, and one of 60 Perception Test evaluations at 20\% loses 15.95\%.  A
smaller pilot may therefore be feasible, but the prespecified 50\% pilot
remains the policy until a reduced pilot is validated on a new target.

The lower chart in Figure~\ref{fig:additional-efficiency} also
clarifies where the signal originates.  Three base responses from different model families
(three calls, 48 model-input frames) give mean gains of 14.11\% and 18.29\%
on HERBench and Perception Test.  Four temporal views from the model being evaluated
alone give only 1.68\% and 1.74\%.  Adding the dense view for all three model
families requires six calls and 120 model-input frames and gives 16.51\% and
19.72\%, close to the full twelve-call response set; because this configuration was identified after the
final-test labels became available, it remains a candidate for a future
prespecified evaluation.

\begin{figure*}[t]
\centering
\includegraphics[width=0.96\textwidth]{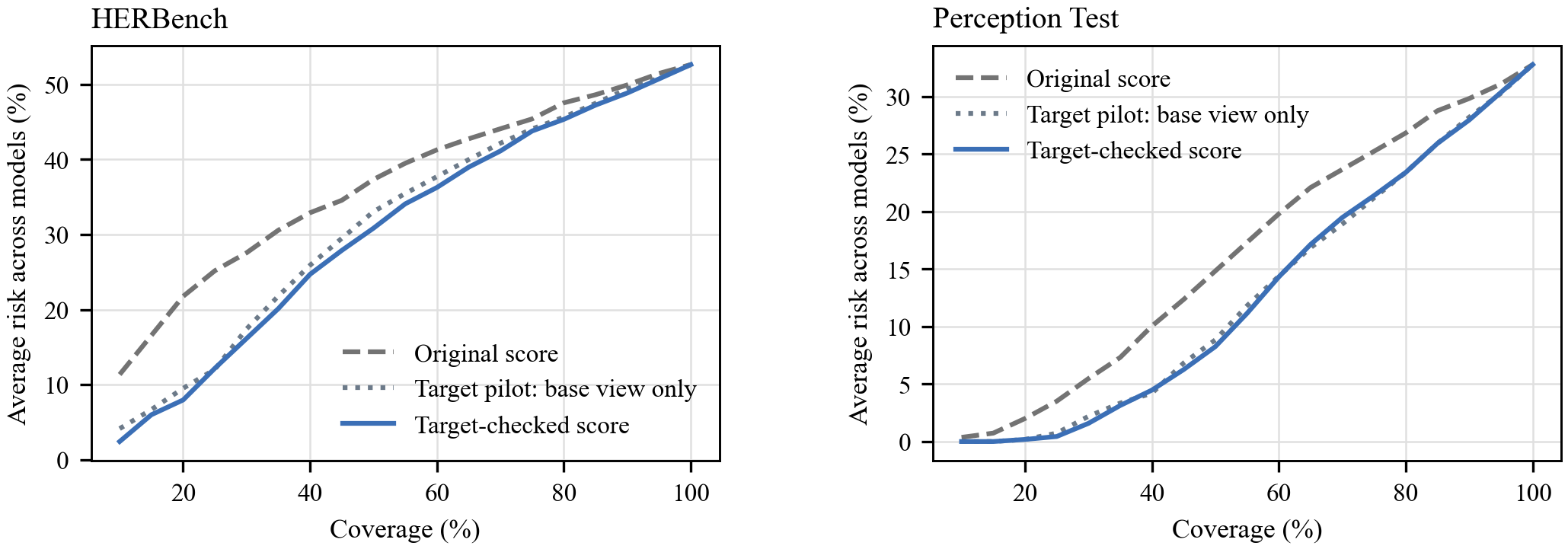}
\caption{Risk-coverage curves on the two held-out test splits, averaged
across the three models.}
\label{fig:risk-coverage}
\end{figure*}

Figure~\ref{fig:risk-coverage} shows what the ranking change means in
practice.  At 50\% coverage, the error rate averaged across the three models
falls from 37.34\% to
30.84\% on HERBench and from 14.84\% to 8.26\% on Perception Test.  At full
coverage all methods have identical risk because the method never changes an
answer.

\subsection{Repeated target splits}
\label{sec:repeated-splits}

To test sensitivity to the split, we inferred a new response bank with the same models and views and repeated the target check on 30 further random video splits per target, with 34 pilot and 34 test videos on HERBench and 250 and 250 on Perception Test (Table~\ref{tab:repeated-splits}).  With its different prompt and runtime, the new bank gives 12.34\% and 15.67\% on the reference splits, against 16.64\% and 18.87\% above.  Over the 30 splits, the target-checked score reduces AURC by 13.66\% and 14.87\%, and none of the 180 model and split evaluations is worse than the original score.  On Perception Test, InternVL keeps its original score in 27 splits.

Training-free support behaves differently on the two targets.  Pooled over all views, it gains 15.27\% on HERBench, more than the checked score in 28 of 30 splits, but on Perception Test it ranks InternVL's answers worse than the original score in 17 splits.  When all-view support was added afterwards as a third candidate, the check chose it in 78 of 90 HERBench evaluations and never for InternVL on Perception Test, giving 14.84\% and 14.45\% with no harm.  Appendix~\ref{app:repeated} gives per-model results and majority voting.

The check matters most with small pilots.  With pilots of 10\%, 20\%, and 30\% of the videos (Table~\ref{tab:small-pilot}), the unchecked pilot score harmed 20, 13, and 4 of 90 Perception Test evaluations, mostly for InternVL3.5, and the checked score 3, 2, and 0.  On HERBench neither harmed any evaluation.

\begin{table}[t]
\color{black}
\centering
\caption{Harmful evaluations (of 90) over the 30 further splits of each target
when the pilot is reduced: evaluations whose test AURC is worse than that of the
original score.  Pilot videos are drawn from each split's pilot half; the test
half is unchanged.  Unchecked always uses the target-pilot logistic score.}
\label{tab:small-pilot}
\PaperTableSetup
\begin{tabularx}{\columnwidth}{@{\PaperTableEdge}Yrrr@{\PaperTableEdge}}
\toprule
Target & Pilot videos & Checked & Unchecked \\
\midrule
HERBench & 7 (10\%) & 0 & 0 \\
HERBench & 14 (20\%) & 0 & 0 \\
HERBench & 20 (30\%) & 0 & 0 \\
HERBench & 34 (50\%) & 0 & 0 \\
Perception & 50 (10\%) & 3 & 20 \\
Perception & 100 (20\%) & 2 & 13 \\
Perception & 150 (30\%) & 0 & 4 \\
Perception & 250 (50\%) & 0 & 0 \\
\bottomrule
\end{tabularx}
\end{table}

We also reconstructed the PhaseGuard features and score selector \cite{liu2026phaseguard} and compared them with ours on the same Perception Test splits, ranking Qwen3-VL-8B base-view answers.  With the same logistic regression, our features gain 18.69\% and PhaseGuard's 18.07\%; the full pipelines gain 17.75\% and 18.15\%, and neither harms any split (Appendix~\ref{app:phaseguard}).

\begin{table}[t]
\color{black}
\centering
\caption{Mean relative AURC reductions (\%) over 30 further video splits of
each target with a newly inferred response bank. Harm counts the 90 model and
split evaluations per target in which a score ranks answers worse than the
original score.}
\label{tab:repeated-splits}
\PaperTableSetup
\begin{tabularx}{\columnwidth}{@{\PaperTableEdge}Yrrrr@{\PaperTableEdge}}
\toprule
& \multicolumn{2}{c}{HERBench} & \multicolumn{2}{c}{Perception Test} \\
\cmidrule(lr){2-3}\cmidrule(lr){4-5}
Score & Mean & Harm & Mean & Harm \\
\midrule
All-view support, training-free & 15.27 & 0 & 11.84 & 17 \\
Base-view support, training-free & 12.31 & 0 & 12.53 & 22 \\
Target-checked score & 13.66 & 0 & 14.87 & 0 \\
\bottomrule
\end{tabularx}
\end{table}
\section{Discussion}
\label{sec:discussion}

\subsection{Findings from Video-MME-v2}

The Video-MME-v2 result changes the interpretation of the development
experiments.  A score trained on one benchmark can rank errors poorly on
another, even when the new videos look similar in simple feature summaries.
The revised method asks a narrower question: on a labeled sample of target
videos, does either predefined score improve AURC\@?  The pilot leaves every
answer unchanged and decides only whether a learned reliability score should
replace the original score.

In both held-out tests, the model-level check passes without accepting a
negative model result.  The decision tables show when the learned score is
used; they do not imply that every task improves.  Retaining the original
score prevents a learned score from
changing the ranking when the pilot
evidence is too weak.

All three HERBench model settings improve, but Scene Verification \&
Arrangement becomes worse in the task-level breakdown.  Perception Test shows
a different pattern: Qwen and LLaVA improve, while InternVL keeps its original
score; all four broad task groups have positive gains when
averaged across the three models.  The
target check controls model-level deployment, not every task group.  If a
deployment has stable task labels, each important group should receive its
own pilot check.  When too few labeled videos are available, that group should
retain the original score.

\subsection{What the response graph contributes}

The VHD diagnostic helps explain why repeated confidence alone is not enough.
On questions where the video contradicts common expectations, three model
families can agree with high confidence and still be wrong.  The response graph
keeps the source of each comparison:
agreement between model families, stability after a sampling change, response
to more frames, and sensitivity to reversing temporal order are separate
measurements.  Training-free cross-model support already captures much of the
ranking gain without labels (Section~\ref{sec:two-confirmations}).  The source ablations and training-free
comparisons support two conclusions: comparisons between model families are
the strongest starting signal, and temporal changes can add information when
they are learned and checked on the target pilot.

The method is not a voting method.  It returns the base answer from one model;
peer responses and alternate views affect only its position in the
risk-coverage ranking.  Accuracy stays fixed, so an AURC change measures the
reliability score rather than a changed answer.

The feature set records standard confidence, divergence, and agreement
measures, and training a correctness model with labels from a new domain has
precedents \cite{kamath2020selective,dancette2023improving}.  Our contribution
is evidence about when such a learned score should replace the original score.
The pre-revision rule, which improved Video-MME and EgoSchema, lost 1.22\% in
its worst Video-MME-v2 setting, and on the held-out Perception Test split the
development HGB score worsened ranking for one model by 4.52\%.  The target
check kept the original score whenever the pilot did not support replacement.
Training-free support, by contrast, harmed 17 of 90 Perception Test evaluations in the repeated splits, while the checked scores harmed none.  Shuffling the relation labels
changed the development gain only from 11.61\% to 10.03\%, so the value of the
response graph lies mainly in which responses are compared, above all responses
from other model families, and less in the relation types.

\subsection{Computational cost and deployment limits}

No answering model is retrained, and its answers are never changed.  The pilot labels could instead be used to fine-tune or switch models, but that changes the answers users receive and needs model weights; the method is meant for fixed models, such as a certified component or one whose answers already feed other systems.  Majority voting, which does change answers, did not help either: on the Perception Test reference split it reached 70.80\% accuracy against 72.67\% for InternVL3.5 alone (Appendix~\ref{app:repeated}).  The graph-based reliability
models are small.  The evaluated model set prioritizes broad evidence coverage, using 12 model calls
and 168 model-input frames per question (three model families under four
temporal views).
The pre-revision lower-cost variant reduced the average number of calls on
EgoSchema, but it did not pass the later Video-MME-v2 target-check requirement
and was not used in the two held-out tests.
If reviewers can check only half of the answers, ranking with the target-checked score lowers the error among the unchecked half from 37.34\% to 30.84\% on HERBench and from 14.84\% to 8.26\% on Perception Test.  This costs 12 model calls per question, about 41.6 GPU seconds per question on Perception Test, and a one-time labeled pilot of 34 HERBench or 250 Perception Test videos.  Post hoc results suggest cheaper settings: pilots of 20 to 27 HERBench or 150 to 200 Perception Test videos made no harmful selection, six calls (base and dense views) came close to twelve, and the InternVL3.5 and Qwen3-VL pair with eight calls kept most of the gain (17.5\% and 15.0\% against 18.3\% and 15.1\%; the pair was chosen on the test splits).  These settings need prespecified validation before use.

\section{Limitations}
\label{sec:limitations}

We reserve half of each target sample for a separate final
test.  This leaves 34 independent HERBench test videos, so the interval based
on complete videos is driven by fewer units than the 963-question count
suggests.  The repeated splits reuse the same 68 HERBench and 500 Perception Test videos, so they measure sensitivity to the split rather than independent replication.
Per-question option probabilities of the held-out runs were not kept, so the released scored predictions reproduce all test metrics and selection decisions but not the scores themselves.  Pilot selection in those runs used a tie-grouped trapezoidal AURC, while all test results use the prefix-risk mean; this changes the test intervals by at most 0.02 points, but the pilot decisions could not be recomputed.  The repeated splits use the prefix-risk mean throughout.
Perception Test provides the larger check, with 250 pilot and 250 test videos.

The evaluated scope is multiple-choice VideoQA with three open model families,
four low-frame visual views, and two target datasets.  Free-form answers,
audio, proprietary models, and later deployment changes remain useful settings
for future evaluation.  Perception Test uses public validation labels,
and public training documentation does not state whether the evaluated
checkpoints have seen these videos. An upstream corpus used by LLaVA-OV2
contains Perception Test annotations, but the subset used for this checkpoint
is unspecified.\footnote{See the training description at
\url{https://arxiv.org/abs/2605.25979} and the upstream corpus at
\url{https://huggingface.co/datasets/OpenGVLab/VideoChat-Flash-Training-Data}.}

The four-view target-pilot score is fitted separately for each model family,
allowing it to adapt to different confidence behavior.  A changed model or
scoring interface therefore requires a new target check.  The stricter margin
for the development HGB candidate is likewise a development choice, not a
universal constant.

The method ranks existing answers and does not by itself correct errors that
all models share.  On VHD questions whose video contradicts common expectations,
85.3\% of unanimous answers are wrong, so cross-model agreement then supports
the wrong answer.  Scores transferred to VHD without its labels ranked
these questions worse than the original score (Appendix~\ref{app:vhd-holdout}).
Such cases need labeled target videos and, ideally, evidence beyond model
agreement.  Temporal views of a single model added little on their own (1.68\%
and 1.74\%); in this study they serve as a supplement to cross-model evidence,
which they improved by 2.54 percentage points on HERBench.

VHD provides a controlled diagnostic analysis, while the two public targets
provide the final tests.  The VHD release includes labels, group assignments,
scoring code, and provenance records; media or retrieval instructions remain
subject to the source licences.

\section{Conclusion}
\label{sec:conclusion}

We refine the reliability scores of VideoQA answers under target shift by
recording responses from temporal views and peer models in a structured graph.
A labeled, video-grouped target pilot decides whether to use one of two
predefined graph scores or retain the original score.  Because the answer never
changes, the reported gains measure improved reliability ordering rather than
answer correction.  On the held-out HERBench split, AURC averaged across the
three model families falls by 16.64\% (95\% complete-video bootstrap CI, 12.11
to 22.59\%); the smallest model-level gain is 11.39\%.  On the separate
Perception Test split, the average reduction is 18.87\% (95\% CI, 15.41 to
22.12\%).  Qwen and LLaVA use the target-pilot score on this split, whereas
InternVL retains its original score.
Across the two targets, AUROC and calibration also improve, and the error rate
at 50\% coverage falls by 6.50 and 6.58 percentage points.  On these reference test splits, it also outperforms all seven training-free
confidence and consistency baselines in AURC averaged across models.  Both
held-out datasets meet the prespecified success criteria without pooling
effects across benchmarks.  Over 30 further splits of each target, the
mean reduction is 13.66\% on HERBench and 14.87\% on Perception Test, and no
split ranks answers worse than the original score.
The development history also shows why the target check matters: an
unlabeled transfer rule that looked safe in earlier evaluations failed on
Video-MME-v2.  The method uses option distributions from peer models and
temporal views without retraining the answering models, but requires labeled
target videos and additional forward calls.

\section*{Data and code availability}

Each test archive contains the dataset version and manifest, the video split,
model and view definitions, fitting settings,
selection thresholds, resampling seeds, and the decision rule.  The HERBench
and Perception Test runs produced 23,652 and 22,656 records, one for each
model, question, and view.
Data preparation and model inference ran on \ifanonymous a university
high-performance computing cluster\else the Katana cluster\fi.

The Python code, experiment records, and scripts needed to reproduce the
results are available in the project repository%
\footnote{\AnonURL{https://github.com/sydney-machine-learning/video-hallucination-diagnosis}}.
The VHD data and annotation code are available from Kaggle and GitHub through
the links given above.  These resources document the data splits, model
revisions, video views, answer records, score selection decisions, and analysis
steps for each test.  The repeated-split response banks are released with per-question option probabilities, together with the scripts and outputs of the additional analyses (Section~\ref{sec:limitations}).

\section*{Declaration of competing interest}

The authors declare that they have no known competing financial interests or
personal relationships that could have appeared to influence the work reported
in this paper.

\section*{Funding}

This research did not receive any specific grant from funding agencies in the
public, commercial, or not-for-profit sectors.
\section*{Declaration of  AI-assisted technologies}

During the preparation of this work, the authors used OpenAI Codex and Anthropic Claude to support
language editing, manuscript organization, literature-search assistance, and
citation and reference consistency checks; the authors reviewed and edited all
outputs as needed and take full responsibility for the content of the published
article.

\begingroup
\bibliographystyle{elsarticle-num}
\bibliography{example}

@inproceedings{li2024mvbench,
  title={{MVBench}: A Comprehensive Multi-modal Video Understanding Benchmark},
  author={Li, Kunchang and Wang, Yali and He, Yinan and Li, Yizhuo and Wang, Yi and Liu, Yi and Wang, Zun and Xu, Jilan and Chen, Guo and Luo, Ping and Wang, Limin and Qiao, Yu},
  booktitle={2024 IEEE/CVF Conference on Computer Vision and Pattern Recognition (CVPR)},
  pages={22195--22206},
  year={2024},
  doi={10.1109/CVPR52733.2024.02095}
}

@inproceedings{fu2025video,
  title={{Video-MME}: The First-Ever Comprehensive Evaluation Benchmark of Multi-modal {LLMs} in Video Analysis},
  author={Fu, Chaoyou and Dai, Yuhan and Luo, Yongdong and Li, Lei and Ren, Shuhuai and Zhang, Renrui and Wang, Zihan and Zhou, Chenyu and Shen, Yunhang and Zhang, Mengdan and Chen, Peixian and Li, Yanwei and Lin, Shaohui and Zhao, Sirui and Li, Ke and Xu, Tong and Zheng, Xiawu and Chen, Enhong and Shan, Caifeng and He, Ran and Sun, Xing},
  booktitle={2025 IEEE/CVF Conference on Computer Vision and Pattern Recognition (CVPR)},
  pages={24108--24118},
  year={2025},
  organization={IEEE},
  doi={10.1109/CVPR52734.2025.02245}
}

@misc{fu2026video,
  title={{Video-MME-v2}: Towards the Next Stage in Benchmarks for Comprehensive Video Understanding},
  author={Fu, Chaoyou and Yuan, Haozhi and Dong, Yuhao and Zhang, Yi-Fan and Shen, Yunhang and Hu, Xiaoxing and Li, Xueying and Su, Jinsen and Long, Chengwu and Xie, Xiaoyao and Xie, Yongkang and Zheng, Xiawu and Yang, Xue and Cao, Haoyu and Wu, Yunsheng and Liu, Ziwei and Sun, Xing and Shan, Caifeng and He, Ran},
  year={2026},
  howpublished={arXiv preprint arXiv:2604.05015},
  doi={10.48550/arXiv.2604.05015},
  url={https://arxiv.org/abs/2604.05015},
  eprint={2604.05015},
  archivePrefix={arXiv}
}

@article{mangalam2023egoschema,
  title={{EgoSchema}: A Diagnostic Benchmark for Very Long-form Video Language Understanding},
  author={Mangalam, Karttikeya and Akshulakov, Raiymbek and Malik, Jitendra},
  journal={Advances in Neural Information Processing Systems},
  volume={36},
  pages={46212--46244},
  year={2023},
  doi={10.52202/075280-2004}
}

@misc{wang2024videohallucer,
  title={{VideoHallucer}: Evaluating Intrinsic and Extrinsic Hallucinations in Large Video-Language Models},
  author={Wang, Yuxuan and Wang, Yueqian and Zhao, Dongyan and Xie, Cihang and Zheng, Zilong},
  year={2024},
  howpublished={arXiv preprint arXiv:2406.16338},
  doi={10.48550/arXiv.2406.16338},
  url={https://arxiv.org/abs/2406.16338},
  eprint={2406.16338},
  archivePrefix={arXiv}
}

@inproceedings{li2025vidhalluc,
  title={{VidHalluc}: Evaluating Temporal Hallucinations in Multimodal Large Language Models for Video Understanding},
  author={Li, Chaoyu and Im, Eun Woo and Fazli, Pooyan},
  booktitle={2025 IEEE/CVF Conference on Computer Vision and Pattern Recognition (CVPR)},
  pages={13723--13733},
  year={2025},
  organization={IEEE},
  doi={10.1109/CVPR52734.2025.01281}
}

@inproceedings{guo2017calibration,
  title={On Calibration of Modern Neural Networks},
  author={Guo, Chuan and Pleiss, Geoff and Sun, Yu and Weinberger, Kilian Q},
  booktitle={Proceedings of the 34th International Conference on Machine Learning},
  volume={70},
  series={Proceedings of Machine Learning Research},
  pages={1321--1330},
  year={2017},
  publisher={PMLR},
  url={https://proceedings.mlr.press/v70/guo17a.html}
}

@inproceedings{ovadia2019can,
  title={Can You Trust Your Model's Uncertainty? {E}valuating Predictive Uncertainty under Dataset Shift},
  author={Ovadia, Yaniv and Fertig, Emily and Ren, Jie and Nado, Zachary and Sculley, David and Nowozin, Sebastian and Dillon, Joshua and Lakshminarayanan, Balaji and Snoek, Jasper},
  booktitle={Advances in Neural Information Processing Systems},
  volume={32},
  pages={13991--14002},
  year={2019},
  url={https://proceedings.neurips.cc/paper/2019/hash/8558cb408c1d76621371888657d2eb1d-Abstract.html},
}

@inproceedings{tu2024empirical,
  title={An Empirical Study into What Matters for Calibrating Vision-Language Models},
  author={Tu, Weijie and Deng, Weijian and Campbell, Dylan and Gould, Stephen and Gedeon, Tom},
  booktitle={Proceedings of the 41st International Conference on Machine Learning},
  volume={235},
  series={Proceedings of Machine Learning Research},
  pages={48791--48808},
  year={2024},
  publisher={PMLR},
  url={https://proceedings.mlr.press/v235/tu24a.html}
}

@inproceedings{lafon2025vilu,
  title={{ViLU}: Learning Vision-Language Uncertainties for Failure Prediction},
  author={Lafon, Marc and Karmim, Yannis and Silva-Rodr{\'\i}guez, Julio and Couairon, Paul and Rambour, Cl{\'e}ment and Fournier-Sniehotta, Rapha{\"e}l and Ben Ayed, Ismail and Dolz, Jose and Thome, Nicolas},
  booktitle={2025 IEEE/CVF International Conference on Computer Vision (ICCV)},
  pages={17807--17817},
  year={2025},
  doi={10.1109/ICCV51701.2025.01654}
}

@inproceedings{schmalfuss2025parc,
  title={{PARC}: A Quantitative Framework Uncovering the Symmetries within Vision Language Models},
  author={Schmalfuss, Jenny and Chang, Nadine and VS, Vibashan and Shen, Maying and Bruhn, Andres and Alvarez, Jose M},
  booktitle={2025 IEEE/CVF Conference on Computer Vision and Pattern Recognition (CVPR)},
  pages={25081--25091},
  year={2025},
  doi={10.1109/CVPR52734.2025.02335}
}

@misc{huang2026counterfactual,
  title={Counterfactual Graph for Multi-Agent {LLM} Calibration},
  author={Huang, Jiatan and Li, Mingchen and Li, Ziming and Kwon, Sunjae and Yu, Hong and Zhang, Chuxu},
  year={2026},
  howpublished={arXiv preprint arXiv:2605.30653},
  doi={10.48550/arXiv.2605.30653},
  url={https://arxiv.org/abs/2605.30653},
  eprint={2605.30653},
  archivePrefix={arXiv}
}

@inproceedings{hamidieh2026complementing,
  title={Complementing Self-Consistency with Cross-Model Disagreement for Uncertainty Quantification},
  author={Hamidieh, Kimia and Thost, Veronika and Gerych, Walter and Yurochkin, Mikhail and Ghassemi, Marzyeh},
  booktitle={International Conference on Learning Representations},
  year={2026},
  url={https://proceedings.iclr.cc/paper_files/paper/2026/hash/83e9c3ea5d0b5de331708058637ada10-Abstract-Conference.html}
}

@misc{gorbett2026cross,
  title={Cross-Model Disagreement as a Label-Free Correctness Signal},
  author={Gorbett, Matt and Jana, Suman},
  year={2026},
  howpublished={arXiv preprint arXiv:2603.25450},
  doi={10.48550/arXiv.2603.25450},
  url={https://arxiv.org/abs/2603.25450},
  eprint={2603.25450},
  archivePrefix={arXiv}
}

@misc{kallem2026learning,
  title={Learning to Trust the Crowd: A Multi-Model Consensus Reasoning Engine for Large Language Models},
  author={Kallem, Pranav},
  year={2026},
  howpublished={arXiv preprint arXiv:2601.07245},
  doi={10.48550/arXiv.2601.07245},
  url={https://arxiv.org/abs/2601.07245},
  eprint={2601.07245},
  archivePrefix={arXiv}
}

@inproceedings{geifman2017selective,
  title={Selective Classification for Deep Neural Networks},
  author={Geifman, Yonatan and El-Yaniv, Ran},
  booktitle={Advances in Neural Information Processing Systems},
  volume={30},
  pages={4878--4887},
  year={2017},
  url={https://proceedings.neurips.cc/paper/2017/hash/4a8423d5e91fda00bb7e46540e2b0cf1-Abstract.html},
}

@inproceedings{zhou2025novel,
  title={A Novel Characterization of the Population Area Under the Risk Coverage Curve ({AURC}) and Rates of Finite Sample Estimators},
  author={Zhou, Han and Van Landeghem, Jordy and Popordanoska, Teodora and Blaschko, Matthew B},
  booktitle={Proceedings of the 42nd International Conference on Machine Learning},
  volume={267},
  series={Proceedings of Machine Learning Research},
  pages={79226--79253},
  year={2025},
  publisher={PMLR},
  url={https://proceedings.mlr.press/v267/zhou25y.html}
}

@inproceedings{khan2024consistency,
  title={Consistency and Uncertainty: Identifying Unreliable Responses from Black-Box Vision-Language Models for Selective Visual Question Answering},
  author={Khan, Zaid and Fu, Yun},
  booktitle={2024 IEEE/CVF Conference on Computer Vision and Pattern Recognition (CVPR)},
  pages={10854--10863},
  year={2024},
  organization={IEEE},
  doi={10.1109/CVPR52733.2024.01032}
}

@misc{gautam2026videohedge,
  title={{VideoHEDGE}: Entropy-Based Hallucination Detection for Video-{VLMs} via Semantic Clustering and Spatiotemporal Perturbations},
  author={Gautam, Sushant and Midoglu, Cise and Thambawita, Vajira and Riegler, Michael A and Halvorsen, P{\aa}l},
  year={2026},
  howpublished={arXiv preprint arXiv:2601.08557},
  doi={10.48550/arXiv.2601.08557},
  url={https://arxiv.org/abs/2601.08557},
  eprint={2601.08557},
  archivePrefix={arXiv}
}

@inproceedings{xiao2024can,
  title={Can {I} Trust Your Answer? {V}isually Grounded Video Question Answering},
  author={Xiao, Junbin and Yao, Angela and Li, Yicong and Chua, Tat-Seng},
  booktitle={2024 IEEE/CVF Conference on Computer Vision and Pattern Recognition (CVPR)},
  pages={13204--13214},
  year={2024},
  organization={IEEE},
  doi={10.1109/CVPR52733.2024.01254}
}

@inproceedings{wu2021star,
  title={{STAR}: A Benchmark for Situated Reasoning in Real-World Videos},
  author={Wu, Bo and Yu, Shoubin and Chen, Zhenfang and Tenenbaum, Joshua B and Gan, Chuang},
  booktitle={Advances in Neural Information Processing Systems},
  volume={34},
  pages={1981--1993},
  year={2021},
  url={https://bobbywu.com/STAR/},
}

@inproceedings{liu2024tempcompass,
  title={{TempCompass}: Do Video {LLMs} Really Understand Videos?},
  author={Liu, Yuanxin and Li, Shicheng and Liu, Yi and Wang, Yuxiang and Ren, Shuhuai and Li, Lei and Chen, Sishuo and Sun, Xu and Hou, Lu},
  booktitle={Findings of the Association for Computational Linguistics: ACL 2024},
  pages={8731--8772},
  year={2024},
  doi={10.18653/v1/2024.findings-acl.517}
}

@misc{bai2025qwen3,
  title={{Qwen3-VL} Technical Report},
  author={Shuai Bai and Yuxuan Cai and Ruizhe Chen and Keqin Chen and Xionghui Chen and Zesen Cheng and Lianghao Deng and Wei Ding and Chang Gao and Chunjiang Ge and Wenbin Ge and Zhifang Guo and Qidong Huang and Jie Huang and Fei Huang and Binyuan Hui and Shutong Jiang and Zhaohai Li and Mingsheng Li and Mei Li and Kaixin Li and Zicheng Lin and Junyang Lin and Xuejing Liu and Jiawei Liu and Chenglong Liu and Yang Liu and Dayiheng Liu and Shixuan Liu and Dunjie Lu and Ruilin Luo and Chenxu Lv and Rui Men and Lingchen Meng and Xuancheng Ren and Xingzhang Ren and Sibo Song and Yuchong Sun and Jun Tang and Jianhong Tu and Jianqiang Wan and Peng Wang and Pengfei Wang and Qiuyue Wang and Yuxuan Wang and Tianbao Xie and Yiheng Xu and Haiyang Xu and Jin Xu and Zhibo Yang and Mingkun Yang and Jianxin Yang and An Yang and Bowen Yu and Fei Zhang and Hang Zhang and Xi Zhang and Bo Zheng and Humen Zhong and Jingren Zhou and Fan Zhou and Jing Zhou and Yuanzhi Zhu and Ke Zhu},
  year={2025},
  howpublished={arXiv preprint arXiv:2511.21631},
  doi={10.48550/arXiv.2511.21631},
  url={https://arxiv.org/abs/2511.21631},
  eprint={2511.21631},
  archivePrefix={arXiv}
}

@misc{wang2025internvl3,
  title={{InternVL3.5}: Advancing Open-Source Multimodal Models in Versatility, Reasoning, and Efficiency},
  author={Weiyun Wang and Zhangwei Gao and Lixin Gu and Hengjun Pu and Long Cui and Xingguang Wei and Zhaoyang Liu and Linglin Jing and Shenglong Ye and Jie Shao and Zhaokai Wang and Zhe Chen and Hongjie Zhang and Ganlin Yang and Haomin Wang and Qi Wei and Jinhui Yin and Wenhao Li and Erfei Cui and Guanzhou Chen and Zichen Ding and Changyao Tian and Zhenyu Wu and Jingjing Xie and Zehao Li and Bowen Yang and Yuchen Duan and Xuehui Wang and Zhi Hou and Haoran Hao and Tianyi Zhang and Songze Li and Xiangyu Zhao and Haodong Duan and Nianchen Deng and Bin Fu and Yinan He and Yi Wang and Conghui He and Botian Shi and Junjun He and Yingtong Xiong and Han Lv and Lijun Wu and Wenqi Shao and Kaipeng Zhang and Huipeng Deng and Biqing Qi and Jiaye Ge and Qipeng Guo and Wenwei Zhang and Songyang Zhang and Maosong Cao and Junyao Lin and Kexian Tang and Jianfei Gao and Haian Huang and Yuzhe Gu and Chengqi Lyu and Huanze Tang and Rui Wang and Haijun Lv and Wanli Ouyang and Limin Wang and Min Dou and Xizhou Zhu and Tong Lu and Dahua Lin and Jifeng Dai and Weijie Su and Bowen Zhou and Kai Chen and Yu Qiao and Wenhai Wang and Gen Luo},
  year={2025},
  howpublished={arXiv preprint arXiv:2508.18265},
  doi={10.48550/arXiv.2508.18265},
  url={https://arxiv.org/abs/2508.18265},
  eprint={2508.18265},
  archivePrefix={arXiv}
}

@misc{an2026llava,
  title={{LLaVA-OneVision-2}: Towards Next-Generation Perceptual Intelligence},
  author={Xiang An and Yin Xie and Feilong Tang and Yunyao Yan and Huajie Tan and Didi Zhu and Changrui Chen and Xiuwei Zhao and Bin Qin and Kaicheng Yang and Yifei Shen and Yuanhan Zhang and Kaichen Zhang and Wenkang Zhang and Zheng Cheng and Nansen Zhang and Chunsheng Wu and Chunjiang Ge and Zimin Ran and Dehua Song and Chunyuan Li and Shikun Feng and Ming Hu and Zhangquan Chen and Junbo Niu and Bo Li and Ziyong Feng and Ziwei Liu and Zongyuan Ge and Jiankang Deng},
  year={2026},
  howpublished={arXiv preprint arXiv:2605.25979},
  doi={10.48550/arXiv.2605.25979},
  url={https://arxiv.org/abs/2605.25979},
  eprint={2605.25979},
  archivePrefix={arXiv}
}

@inproceedings{srinivasan2024selective,
  title={Selective ``Selective Prediction'': Reducing Unnecessary Abstention in Vision-Language Reasoning},
  author={Srinivasan, Tejas and Hessel, Jack and Gupta, Tanmay and Lin, Bill Yuchen and Choi, Yejin and Thomason, Jesse and Chandu, Khyathi},
  booktitle={Findings of the Association for Computational Linguistics: ACL 2024},
  pages={12935--12948},
  year={2024},
  doi={10.18653/v1/2024.findings-acl.767}
}

@inproceedings{cattelan2024fix,
  title={How to Fix a Broken Confidence Estimator: Evaluating Post-hoc Methods for Selective Classification with Deep Neural Networks},
  author={Cattelan, Lu{\'\i}s Felipe P and Silva, Danilo},
  booktitle={Proceedings of the Fortieth Conference on Uncertainty in Artificial Intelligence},
  volume={244},
  series={Proceedings of Machine Learning Research},
  pages={547--584},
  year={2024},
  publisher={PMLR},
  url={https://proceedings.mlr.press/v244/cattelan24a.html}
}

@inproceedings{park2020calibrated,
  title={Calibrated Prediction with Covariate Shift via Unsupervised Domain Adaptation},
  author={Park, Sangdon and Bastani, Osbert and Weimer, James and Lee, Insup},
  booktitle={Proceedings of the Twenty Third International Conference on Artificial Intelligence and Statistics},
  volume={108},
  series={Proceedings of Machine Learning Research},
  pages={3219--3229},
  year={2020},
  publisher={PMLR},
  url={https://proceedings.mlr.press/v108/park20b.html}
}

@inproceedings{jung2025consistency,
  title={On the Consistency of Video Large Language Models in Temporal Comprehension},
  author={Jung, Minjoon and Xiao, Junbin and Zhang, Byoung-Tak and Yao, Angela},
  booktitle={2025 IEEE/CVF Conference on Computer Vision and Pattern Recognition (CVPR)},
  pages={13713--13722},
  year={2025},
  organization={IEEE},
  doi={10.1109/CVPR52734.2025.01280}
}

@inproceedings{wang2025sample,
  title={Sample then Identify: A General Framework for Risk Control and Assessment in Multimodal Large Language Models},
  author={Wang, Qingni and Geng, Tiantian and Wang, Zhiyuan and Wang, Teng and Fu, Bo and Zheng, Feng},
  booktitle={International Conference on Learning Representations},
  year={2025},
  url={https://proceedings.iclr.cc/paper_files/paper/2025/hash/a1722a6bd1023c026a3d6a570fb3af75-Abstract-Conference.html},
}

@misc{tran2026knowing,
  title={Knowing When to Answer: Adaptive Confidence Refinement for Reliable Audio-Visual Question Answering},
  author={Tran, Dinh Phu and Jeong, Jihoon and Wazir, Saad and Kim, Seongah and Do, Thao and Subakan, Cem and Kim, Daeyoung},
  year={2026},
  howpublished={arXiv preprint arXiv:2602.04924},
  doi={10.48550/arXiv.2602.04924},
  url={https://arxiv.org/abs/2602.04924},
  eprint={2602.04924},
  archivePrefix={arXiv}
}

@inproceedings{ben2026herbench,
  title={{HERBench}: A Benchmark for Multi-Evidence Integration in Video Question Answering},
  author={Ben-Ami, Dan and Serussi, Gabriele and Cohen, Kobi and Baskin, Chaim},
  booktitle={2026 IEEE/CVF Conference on Computer Vision and Pattern Recognition (CVPR)},
  pages={4505--4514},
  year={2026},
  url={https://openaccess.thecvf.com/content/CVPR2026/html/Ami_HERBench_A_Benchmark_for_Multi-Evidence_Integration_in_Video_Question_Answering_CVPR_2026_paper.html}
}

@article{angelopoulos2025learn,
  title={Learn then Test: Calibrating Predictive Algorithms to Achieve Risk Control},
  author={Angelopoulos, Anastasios N and Bates, Stephen and Cand{\`e}s, Emmanuel J and Jordan, Michael I and Lei, Lihua},
  journal={The Annals of Applied Statistics},
  volume={19},
  number={2},
  pages={1641--1662},
  year={2025},
  publisher={Institute of Mathematical Statistics},
  doi={10.1214/24-AOAS1998}
}

@inproceedings{almeida2025high,
  title={High Probability Risk Control Under Covariate Shift},
  author={Almeida, Duarte C and Bravo, Jo{\~a}o and Bono, Jacopo and Bizarro, Pedro and Figueiredo, M{\'a}rio AT},
  booktitle={Proceedings of the Fourteenth Symposium on Conformal and Probabilistic Prediction with Applications},
  volume={266},
  series={Proceedings of Machine Learning Research},
  pages={133--152},
  year={2025},
  publisher={PMLR},
  url={https://proceedings.mlr.press/v266/almeida25a.html}
}

@article{wieczorek2026variational,
  title={Variational Visual Question Answering for Uncertainty-Aware Selective Prediction},
  author={Wieczorek, Tobias Jan and Daun, Nathalie and Khan, Mohammad Emtiyaz and Rohrbach, Marcus},
  journal={Transactions on Machine Learning Research},
  year={2026},
  url={https://openreview.net/forum?id=jtnMIbJIso},
}

@inproceedings{wang2026coin,
  title={{COIN}: Uncertainty-Guarding Selective Question Answering for Foundation Models with Provable Risk Guarantees},
  author={Wang, Zhiyuan and Duan, Jinhao and Wang, Qingni and Zhu, Xiaofeng and Chen, Tianlong and Shi, Xiaoshuang and Xu, Kaidi},
  booktitle={Proceedings of the AAAI Conference on Artificial Intelligence},
  volume={40},
  pages={33764--33772},
  year={2026},
  doi={10.1609/aaai.v40i40.40667}
}

@misc{dong2026uncertainty,
  title={Uncertainty-Aware Abstention in Large Language Models with Provable Alignment Guarantees},
  author={Dong, Sijin and Shinnou, Hiroyuki},
  year={2026},
  howpublished={arXiv preprint arXiv:2607.04430},
  doi={10.48550/arXiv.2607.04430},
  url={https://arxiv.org/abs/2607.04430},
  eprint={2607.04430},
  archivePrefix={arXiv}
}

@inproceedings{kamath2020selective,
  title={Selective Question Answering under Domain Shift},
  author={Kamath, Amita and Jia, Robin and Liang, Percy},
  booktitle={Proceedings of the 58th Annual Meeting of the Association for Computational Linguistics},
  pages={5684--5696},
  year={2020},
  doi={10.18653/v1/2020.acl-main.503}
}

@inproceedings{whitehead2022reliable,
  title={Reliable Visual Question Answering: Abstain Rather Than Answer Incorrectly},
  author={Whitehead, Spencer and Petryk, Suzanne and Shakib, Vedaad and Gonzalez, Joseph and Darrell, Trevor and Rohrbach, Anna and Rohrbach, Marcus},
  booktitle={European Conference on Computer Vision},
  pages={148--166},
  year={2022},
  publisher={Springer},
  doi={10.1007/978-3-031-20059-5_9}
}

@inproceedings{dancette2023improving,
  title={Improving Selective Visual Question Answering by Learning from Your Peers},
  author={Dancette, Corentin and Whitehead, Spencer and Maheshwary, Rishabh and Vedantam, Ramakrishna and Scherer, Stefan and Chen, Xinlei and Cord, Matthieu and Rohrbach, Marcus},
  booktitle={2023 IEEE/CVF Conference on Computer Vision and Pattern Recognition (CVPR)},
  pages={24049--24059},
  year={2023},
  organization={IEEE},
  doi={10.1109/CVPR52729.2023.02303}
}

@article{patraucean2023perception,
  title={{Perception Test}: A Diagnostic Benchmark for Multimodal Video Models},
  author={Patraucean, Viorica and Smaira, Lucas and Gupta, Ankush and Recasens, Adria and Markeeva, Larisa and Banarse, Dylan and Koppula, Skanda and Heyward, Joseph and Malinowski, Mateusz and Yang, Yi and Doersch, Carl and Matejovicova, Tatiana and Sulsky, Yury and Miech, Antoine and Fr{\'e}chette, Alexandre and Klimczak, Hanna and Koster, Raphael and Zhang, Junlin and Winkler, Stephanie and Aytar, Yusuf and Osindero, Simon and Damen, Dima and Zisserman, Andrew and Carreira, Joao},
  journal={Advances in Neural Information Processing Systems},
  volume={36},
  pages={42748--42761},
  year={2023},
  doi={10.52202/075280-1852}
}

@inproceedings{bae2025mash,
  title={{MASH-VLM}: Mitigating Action-Scene Hallucination in Video-{LLMs} through Disentangled Spatial-Temporal Representations},
  author={Bae, Kyungho and Kim, Jinhyung and Lee, Sihaeng and Lee, Soonyoung and Lee, Gunhee and Choi, Jinwoo},
  booktitle={2025 IEEE/CVF Conference on Computer Vision and Pattern Recognition (CVPR)},
  pages={13744--13753},
  year={2025},
  publisher={IEEE},
  doi={10.1109/CVPR52734.2025.01283}
}

@inproceedings{liu2026phaseguard,
  title={Learning When to Trust and Reobserve Sparse-Frame {VideoQA} from Temporal Sampling Perturbations},
  author={Liu, Shujia},
  year={2026},
  booktitle={Advances in Neural Information Processing Systems},
  eprint={2610.23969},
  archivePrefix={arXiv},
  doi={10.48550/arXiv.2610.23969},
  url={https://arxiv.org/abs/2610.23969}
}

@misc{liu2026jury,
  title={{LLMs} as a Jury: Cross-Model Consensus Can Outperform Process Reward Models for {LLM} Reasoning},
  author={Liu, Ning},
  year={2026},
  howpublished={arXiv preprint arXiv:2607.10139},
  doi={10.48550/arXiv.2607.10139},
  url={https://arxiv.org/abs/2607.10139}
}

\appendix

\section{Detailed results and notation}
\label{app:detailed-results}

\subsection{Notation and plain-language guide}
\label{app:notation}

\begin{table}[H]
\centering
\caption{Notation used in the method.}
\label{tab:notation}
\PaperTableSetup
\begin{tabular}{@{\PaperTableEdge}lp{0.70\columnwidth}@{\PaperTableEdge}}
\toprule
Symbol & Meaning \\
\midrule
$v,c$ & temporal view and answer option \\
$p_v(c)$ & probability assigned to option $c$ in view $v$ \\
$a,b$ & unchanged base answer and correct option \\
$y$ & 1 when $a=b$, otherwise 0 \\
$G$ & response graph for one question \\
$z$ & numerical measurements calculated from $G$ \\
$r,s$ & original score and the candidate score being checked \\
$g,L$ & relative AURC reduction and its lower bound \\
\bottomrule
\end{tabular}
\end{table}

\label{app:plain-language}

\subsection{Development view definitions}
\label{app:development-views}

Development view definitions and frame budgets vary. NExT-GQA uses 16 frames
per view: uniform and shifted sampling, annotated evidence intervals, and their
complements. The latter two fill the added-evidence and removed-evidence
feature groups. STAR and MVBench each use eight frames for the base, shifted,
and reversed views, and sixteen for the dense view. VHD has base and dense views
with 8 and 16 frames, and TempCompass has the same two views with 6 and 12
frames. Missing groups have zero-filled comparisons and explicit availability
indicators.

\subsection{Feature vector}
\label{app:features}

The reference model is the model whose answer is scored.  On the final
targets, each base answer has one feature vector \(z\) with
\(4+3+4+12+30=53\) entries.  The base response contributes four: confidence,
entropy, the gap between the two largest probabilities, and the number of
options divided by five.  Three indicators identify the reference model.  Four
entries summarize the base view across the three models: vote fraction, mean
support, support standard deviation, and consensus entropy.  Twelve entries
compare individual models: for each of the three models, whether it is the
reference model, whether its answer agrees, its support for \(a\), and its JS
divergence from the reference response.  The last 30 entries come from the
other three views, ten from each: five describe the reference model (response
availability, JS divergence, entropy change, support change, and answer change)
and five summarize all models (fraction with available responses, vote
fraction, mean support, support standard deviation, and consensus entropy).
Missing responses have zero-filled values and explicit availability indicators.

\subsection{Task-level results}
\label{app:herbench-tasks}

Table~\ref{tab:herbench-tasks} reports the final target-checked score under the
model-level decisions made on the pilot.  The rows are descriptive; a separate
method is not selected for each task.

\begin{table}[H]
\centering
\caption{HERBench final-test results by task using condensed task labels.}
\label{tab:herbench-tasks}
\PaperTableSetup
\setlength{\tabcolsep}{3pt}
\begin{tabularx}{\columnwidth}{@{\PaperTableEdge}Yrrrr@{\PaperTableEdge}}
\toprule
Task & Scored rows & Accuracy & Mean gain & Worst \\
\midrule
Action Counting & 195 & 0.708 & 42.97\% & -3.55\% \\
Action Sequence & 216 & 0.634 & 19.74\% & 8.19\% \\
Appearance: Attributes & 45 & 0.778 & 54.13\% & 44.84\% \\
Appearance: Behavior/Interactions & 132 & 0.902 & 26.97\% & 16.11\% \\
Appearance: Localization/Trajectory & 273 & 0.795 & 27.63\% & 14.62\% \\
False Action Memory & 225 & 0.507 & 12.47\% & -13.48\% \\
False Object Memory & 228 & 0.461 & 20.88\% & 6.94\% \\
Multi-Entity Grounding/Localization & 333 & 0.261 & -0.86\% & -1.56\% \\
Multi-Person Duration Reasoning & 285 & 0.361 & 2.96\% & -9.97\% \\
Region People Counting & 234 & 0.338 & 7.60\% & 0.89\% \\
Scene Verification/Arrangement & 492 & 0.388 & -18.60\% & -45.32\% \\
Temporal Ordering & 231 & 0.186 & 8.93\% & 0.00\% \\
\bottomrule
\end{tabularx}
\end{table}

\label{app:perception-tasks}

Table~\ref{tab:perception-tasks} reports the four official area
groups on the held-out split.  Scored rows count pairs of models and questions.  The
decision on which score to use remains fixed at the model level, so these rows describe
the result rather than selecting a different method for each area.

\begin{table}[H]
\centering
\caption{Perception Test final-test results by area.}
\label{tab:perception-tasks}
\PaperTableSetup
\begin{tabularx}{\columnwidth}{@{\PaperTableEdge}Yrrrr@{\PaperTableEdge}}
\toprule
Area & Scored rows & Accuracy & Mean gain & Worst \\
\midrule
abstraction & 855 & 0.561 & 18.29\% & 0.00\% \\
memory & 114 & 0.544 & 15.72\% & 0.00\% \\
physics & 975 & 0.656 & 20.37\% & 0.00\% \\
semantics & 789 & 0.830 & 22.04\% & 0.00\% \\
\bottomrule
\end{tabularx}
\end{table}

\subsection{VHD diagnostic details}
\label{app:vhd-values}

Table~\ref{tab:vhd-annotation-agreement} reports agreement before an
adjudicator resolved differences.  The number of comparable pairs varies
because prior, visual-evidence, and uncertainty labels apply only to their
corresponding question types and only when both reviewers first marked the
question as valid.

\begin{table}[H]
\centering
\caption{Agreement between the two VHD question reviewers.}
\label{tab:vhd-annotation-agreement}
\PaperTableSetup
\begin{tabular}{@{\PaperTableEdge}lrr@{\PaperTableEdge}}
\toprule
Label & Pairs & Agreement / \(\kappa\) \\
\midrule
Question valid & 1,155 & 97.7\% / 0.945 \\
Correct option & 789 & 89.2\% / 0.785 \\
Language prior & 180 & 89.4\% / 0.789 \\
Video evidence & 180 & 89.4\% / 0.789 \\
Uncertainty & 282 & 90.1\% / 0.796 \\
\bottomrule
\end{tabular}
\end{table}

\begin{table}[H]
\centering
\caption{VHD consensus diagnostic over three model families with fixed weights.}
\label{tab:vhd-consensus-diagnostic}
\PaperTableSetup
\begin{tabularx}{\columnwidth}{@{\PaperTableEdge}Yrrrr@{\PaperTableEdge}}
\toprule
Task & $N$ &
\shortstack[r]{All models\\agree (\%)} &
\shortstack[r]{All agree and\\are wrong (\%)} &
\shortstack[r]{Wrong when all agree\\(\%, 95\% CI)} \\
\midrule
Visible evidence & 333 & 74.2 & 23.1 & 31.2 [25.4, 37.0] \\
Video conflict & 178 & 65.2 & 55.6 & 85.3 [78.6, 91.5] \\
Hidden evidence & 290 & 77.2 & 12.8 & 16.5 [11.8, 21.5] \\
\bottomrule
\end{tabularx}
\end{table}

\subsection{VHD annotation interface}
\label{app:vhd-interface}

Figure~\ref{fig:vhd-annotation-interface} shows the three annotation conditions
for one source clip.  These conditions are separate from the four model-input
views used by the reliability method.  The original condition measures
accuracy when evidence is clearly visible; the reversed condition records
whether an option follows language expectation or the video; and the masked
condition records whether the remaining evidence is sufficient.  The
annotation and review software link is provided with the VHD dataset
description in Section~\ref{sec:experiments}.

\begin{figure*}[p]
\centering
\includegraphics[width=0.85\textwidth]{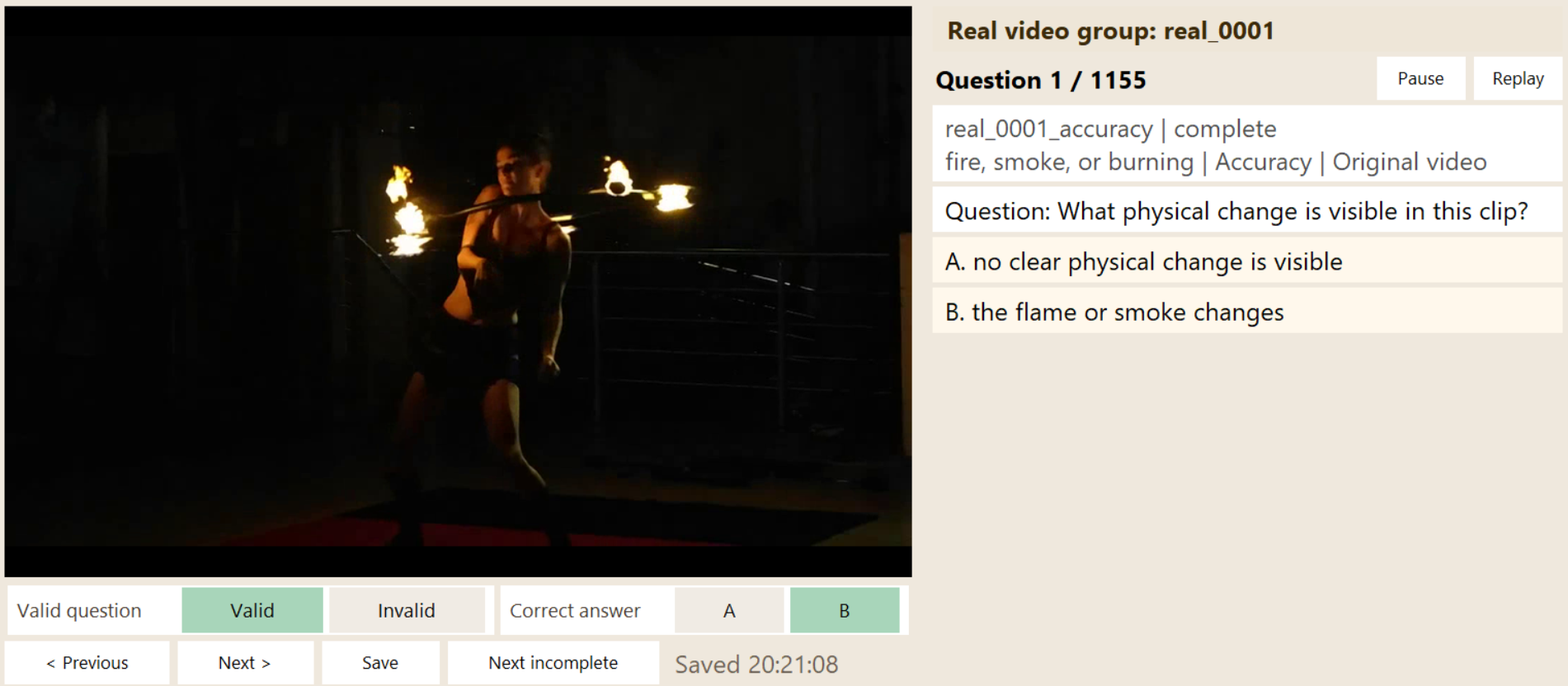}\par
{\normalfont\small (a) Original view\par}
\vspace{1.5mm}

\includegraphics[width=0.85\textwidth]{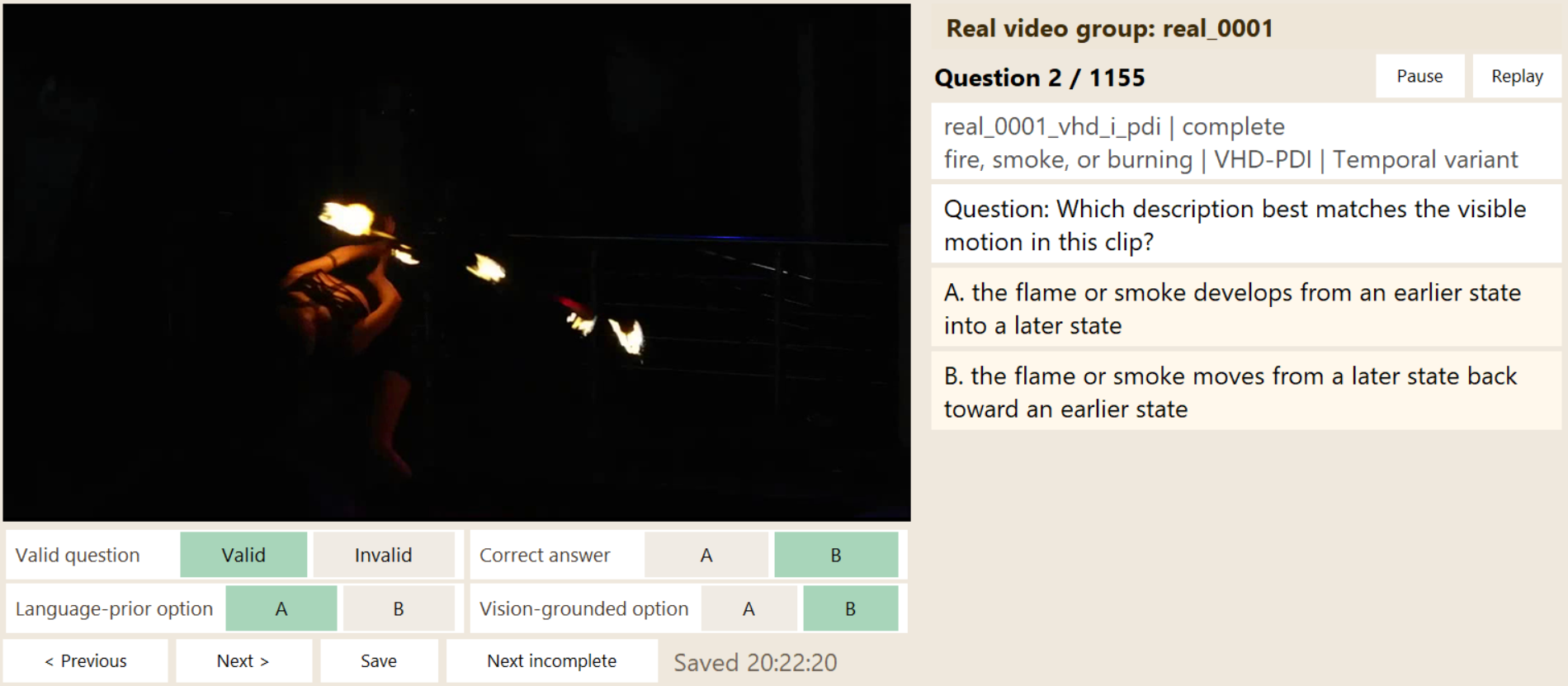}\par
{\normalfont\small (b) Reversed view\par}
\vspace{1.5mm}

\includegraphics[width=0.85\textwidth]{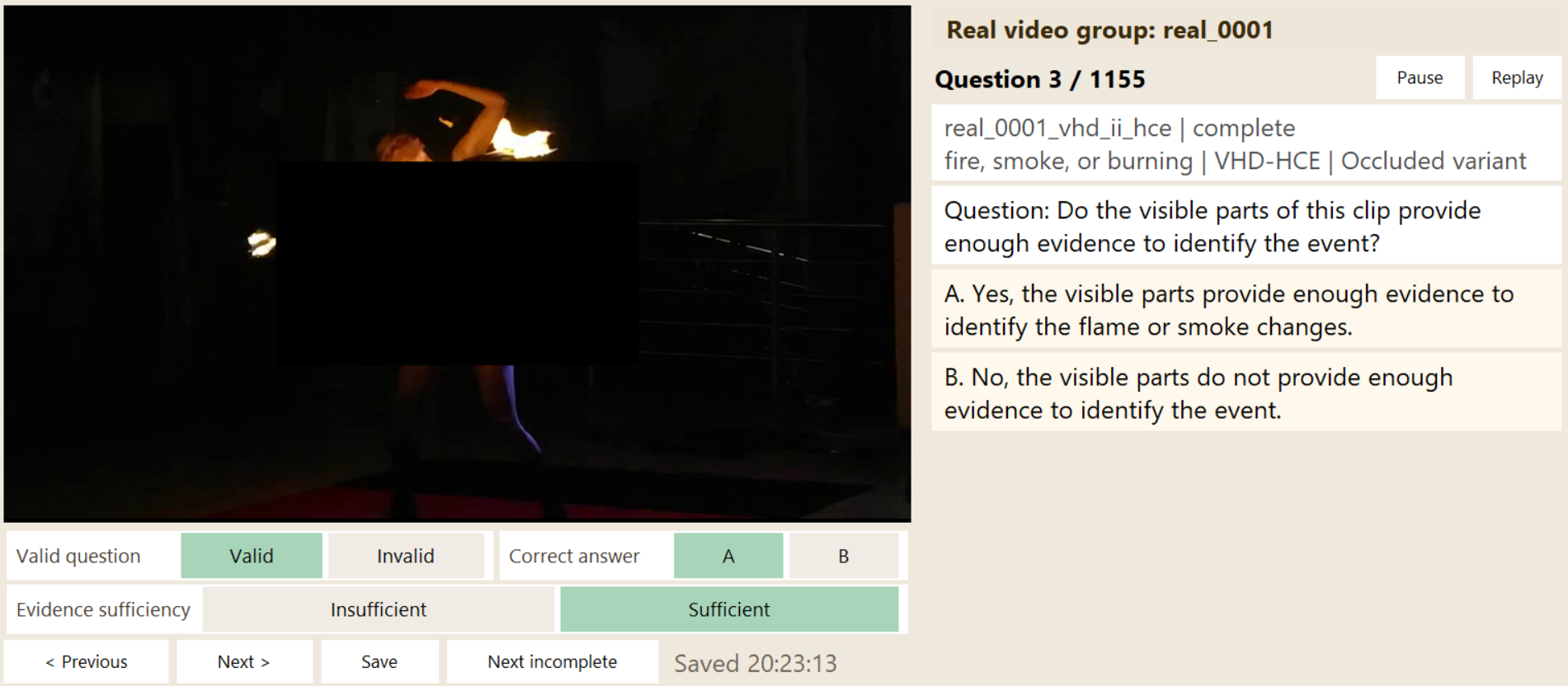}\par
{\normalfont\small (c) Masked view\par}
\caption{VHD annotation conditions for the same source clip.}
\label{fig:vhd-annotation-interface}
\end{figure*}

\section{Supplementary comparisons and checks}
\label{app:robustness-resources}

\subsection{Pilot decisions}
\label{app:pilot-decisions}

Tables~\ref{tab:herbench-decisions} and~\ref{tab:perception-decisions} give,
for each answering model, the score used, the pilot size, the pilot gain, and
the lower bound used by the target check.

\begin{table}[H]
\centering
\caption{HERBench target-pilot score decisions.}
\label{tab:herbench-decisions}
\PaperTableSetup
\begin{tabularx}{\columnwidth}{@{\PaperTableEdge}YYrrrr@{\PaperTableEdge}}
\toprule
Model & Score used & Videos & Qs. & Pilot gain & Lower bound \\
\midrule
InternVL3.5 & target pilot & 34 & 1008 & 12.70\% & 6.78\% \\
Qwen3-VL & target pilot & 34 & 1008 & 31.39\% & 23.20\% \\
LLaVA-OV2 & target pilot & 34 & 1008 & 12.25\% & 4.87\% \\
\bottomrule
\end{tabularx}
\end{table}

\begin{table}[H]
\centering
\caption{Perception Test target-pilot score decisions.}
\label{tab:perception-decisions}
\PaperTableSetup
\begin{tabularx}{\columnwidth}{@{\PaperTableEdge}YYrrrr@{\PaperTableEdge}}
\toprule
Model & Score used & Videos & Qs. & Pilot gain & Lower bound \\
\midrule
InternVL3.5 & original score & 250 & 977 & 0.00\% & 0.00\% \\
Qwen3-VL & target pilot & 250 & 977 & 27.19\% & 21.05\% \\
LLaVA-OV2 & target pilot & 250 & 977 & 29.62\% & 23.91\% \\
\bottomrule
\end{tabularx}
\end{table}

\subsection{Additional baseline comparisons}
\label{app:confirmation-baselines}

Tables~\ref{tab:herbench-baselines} and~\ref{tab:perception-baselines} provide
the values summarized in
Figure~\ref{fig:heldout-overview}(b) and (c).  Mean and minimum are relative
AURC changes from the original score across the three models.

\begin{table}[H]
\centering
\caption{HERBench reliability baselines.}
\label{tab:herbench-baselines}
\PaperTableSetup
\begin{tabularx}{\columnwidth}{@{\PaperTableEdge}Yrrrrr@{\PaperTableEdge}}
\toprule
Method & Mean & Min. & AUROC & ECE & Brier \\
\midrule
Target-checked score & 16.64\% & 11.39\% & 0.799 & 0.047 & 0.180 \\
Development logistic & 16.36\% & 12.18\% & 0.798 & 0.126 & 0.196 \\
Development HGB & 14.33\% & 8.09\% & 0.779 & 0.117 & 0.202 \\
Pilot, base view & 14.11\% & 8.93\% & 0.783 & 0.050 & 0.187 \\
Pilot, confidence & 0.27\% & -0.00\% & 0.717 & 0.049 & 0.212 \\
Original score & 0.00\% & 0.00\% & 0.716 & 0.185 & 0.259 \\
\bottomrule
\end{tabularx}
\end{table}

\begin{table}[H]
\centering
\caption{Perception Test reliability baselines.}
\label{tab:perception-baselines}
\PaperTableSetup
\begin{tabularx}{\columnwidth}{@{\PaperTableEdge}Yrrrrr@{\PaperTableEdge}}
\toprule
Method & Mean & Min. & AUROC & ECE & Brier \\
\midrule
Target-checked score & 18.87\% & 0.00\% & 0.844 & 0.074 & 0.154 \\
Pilot, base view & 18.29\% & -0.86\% & 0.843 & 0.119 & 0.167 \\
Development HGB & 14.98\% & -4.52\% & 0.826 & 0.086 & 0.161 \\
Development logistic & 14.91\% & -3.92\% & 0.830 & 0.088 & 0.164 \\
Original score & 0.00\% & 0.00\% & 0.767 & 0.129 & 0.205 \\
Pilot, confidence & -0.09\% & -0.32\% & 0.767 & 0.145 & 0.204 \\
\bottomrule
\end{tabularx}
\end{table}

Table~\ref{tab:training-free} adds seven scores that require no
training.  ``Score gain'' is the score's average relative AURC reduction from
the original score.  \(\Delta\) is the target-checked method's gain minus that
value, so a positive number favors the method; intervals resample complete
videos.

\begin{table}[H]
\centering
\caption{Training-free confidence and consistency scores.}
\label{tab:training-free}
\PaperTableSetup
\begin{tabularx}{\columnwidth}{@{\PaperTableEdge}lYrrr@{\PaperTableEdge}}
\toprule
Target & Score & Score gain & $\Delta$ (pp) & 95\% CI \\
\midrule
HERBench & Top-two margin & -1.20\% & 17.84 & [13.30, 24.16] \\
HERBench & Negative entropy & -0.08\% & 16.73 & [12.05, 23.15] \\
HERBench & Base vote & 8.91\% & 7.73 & [4.40, 12.21] \\
HERBench & Base support & 13.12\% & 3.53 & [1.35, 6.74] \\
HERBench & Own-view stability & -15.15\% & 31.80 & [24.92, 39.75] \\
HERBench & All-view vote & 12.53\% & 4.11 & [1.09, 7.21] \\
HERBench & All-view support & 15.57\% & 1.08 & [-0.70, 2.91] \\
Perception & Top-two margin & -0.10\% & 18.97 & [15.44, 22.34] \\
Perception & Negative entropy & -0.65\% & 19.52 & [15.80, 23.04] \\
Perception & Base vote & 8.17\% & 10.70 & [6.08, 15.41] \\
Perception & Base support & 11.67\% & 7.20 & [2.76, 11.86] \\
Perception & Own-view stability & -36.09\% & 54.96 & [42.31, 69.76] \\
Perception & All-view vote & 6.80\% & 12.07 & [6.45, 18.39] \\
Perception & All-view support & 11.39\% & 7.48 & [3.04, 12.20] \\
\bottomrule
\end{tabularx}
\end{table}

\label{app:posthoc-paired}

Table~\ref{tab:paired-baselines} reports differences in average relative
AURC reduction between the method and each baseline.  Positive values favor
the method.  The paired video-resampling intervals are descriptive and are not
adjusted across the five baseline rows.  The original-score intervals come from
this separate post-evaluation bootstrap run, so their Monte Carlo endpoints
differ slightly from the prespecified confirmation intervals in the main text.

\begin{table}[H]
\centering
\caption{Paired baseline comparisons on the fixed held-out test splits.}
\label{tab:paired-baselines}
\PaperTableSetup
\begin{tabularx}{\columnwidth}{@{\PaperTableEdge}lYrr@{\PaperTableEdge}}
\toprule
Target & Baseline & Gain (pp) & 95\% CI \\
\midrule
HERBench & Original score & 16.64 & [12.09, 22.54] \\
HERBench & Pilot, confidence & 16.37 & [11.82, 22.36] \\
HERBench & Pilot, base view & 2.54 & [0.34, 5.48] \\
HERBench & Development logistic & 0.28 & [-1.01, 1.35] \\
HERBench & Development HGB & 2.32 & [0.15, 4.84] \\
Perception & Original score & 18.87 & [15.51, 22.08] \\
Perception & Pilot, confidence & 18.96 & [15.51, 22.22] \\
Perception & Pilot, base view & 0.58 & [-1.00, 2.10] \\
Perception & Development logistic & 3.96 & [0.54, 7.53] \\
Perception & Development HGB & 3.89 & [1.41, 6.31] \\
\bottomrule
\end{tabularx}
\end{table}

E-AURC (Table~\ref{tab:eaurc}) is AURC minus the AURC of an oracle ordering that puts every correct answer first.  It removes the effect of accuracy, so models with different accuracy can be compared.

\begin{table}[H]
\color{black}
\centering
\caption{Excess AURC on the held-out test splits.  Oracle is the AURC of the
oracle ordering; Original and Checked give E-AURC for the original and
target-checked scores; Closed is the relative reduction in E-AURC.}
\label{tab:eaurc}
\PaperTableSetup
\begin{tabularx}{\columnwidth}{@{\PaperTableEdge}lYrrrr@{\PaperTableEdge}}
\toprule
Target & Model & Oracle & Original & Checked & Closed \\
\midrule
HERBench & InternVL3.5 & 0.1799 & 0.1478 & 0.1105 & 25.3\% \\
HERBench & Qwen3-VL & 0.1712 & 0.2015 & 0.1073 & 46.7\% \\
HERBench & LLaVA-OV2 & 0.1674 & 0.1479 & 0.1060 & 28.3\% \\
HERBench & Mean & 0.1728 & 0.1657 & 0.1079 & 33.4\% \\
Perception & InternVL3.5 & 0.0411 & 0.0597 & 0.0597 & 0.0\% \\
Perception & Qwen3-VL & 0.0607 & 0.0950 & 0.0512 & 46.1\% \\
Perception & LLaVA-OV2 & 0.0858 & 0.1037 & 0.0497 & 52.0\% \\
Perception & Mean & 0.0626 & 0.0861 & 0.0535 & 32.7\% \\
\bottomrule
\end{tabularx}
\end{table}

\subsection{Pilot size and model-call checks}
\label{app:posthoc-pilot}

Table~\ref{tab:pilot-sensitivity} keeps the original test split fixed and
uses smaller subsets of the original pilot videos.  Fractions below 50\%
use 20 video subsets and three evaluation settings; the 50\% row is the single
original pilot.  Cases that return to the original score have zero test gain
and are included in the mean.

\begin{table}[H]
\centering
\caption{Sensitivity to the number of labeled pilot videos.}
\label{tab:pilot-sensitivity}
\PaperTableSetup
\setlength{\tabcolsep}{1.8pt}
\begin{tabularx}{\columnwidth}{@{\PaperTableEdge}Yrrrrr@{\PaperTableEdge}}
\toprule
Target & Pilot & Videos & \shortstack[r]{Learned\\used} & Gain & \shortstack[r]{Harmful\\use} \\
\midrule
HERBench & 10\% & 7 & 55.0\% & 9.43\% & 1.7\% \\
HERBench & 20\% & 14 & 73.3\% & 11.64\% & 0.0\% \\
HERBench & 30\% & 20 & 85.0\% & 13.76\% & 0.0\% \\
HERBench & 40\% & 27 & 100.0\% & 16.14\% & 0.0\% \\
HERBench & 50\% & 34 & 100.0\% & 16.64\% & 0.0\% \\
Perception & 10\% & 50 & 56.7\% & 13.96\% & 0.0\% \\
Perception & 20\% & 100 & 68.3\% & 16.60\% & 1.7\% \\
Perception & 30\% & 150 & 66.7\% & 17.67\% & 0.0\% \\
Perception & 40\% & 200 & 66.7\% & 18.48\% & 0.0\% \\
Perception & 50\% & 250 & 66.7\% & 18.87\% & 0.0\% \\
\bottomrule
\end{tabularx}
\end{table}

\label{app:posthoc-compute}

Table~\ref{tab:model-input-cost} varies the fixed model set while keeping the
pilot and test split fixed.  These rows evaluate target-pilot logistic scores
without the target check; they are not newly tested target-check policies.
Two-family results average all six ordered pairs of evaluated and peer models,
whereas the remaining rows average over each of the three evaluated models.

\begin{table}[H]
\centering
\caption{Response-set ablation and model-input cost.}
\label{tab:model-input-cost}
\PaperTableSetup
\begin{tabularx}{\columnwidth}{@{\PaperTableEdge}lYrrrr@{\PaperTableEdge}}
\toprule
Target & Model set & Calls & Frames & Mean & Min. \\
\midrule
HERBench & Confidence only & 1 & 16 & 0.27\% & 0.00\% \\
HERBench & Base, 3 families & 3 & 48 & 14.11\% & 8.93\% \\
HERBench & Base+dense, 3 families & 6 & 120 & 16.51\% & 11.81\% \\
HERBench & Own 4 views & 4 & 56 & 1.68\% & -1.00\% \\
HERBench & 4 views, 2 families & 8 & 112 & 12.56\% & 2.74\% \\
HERBench & 4 views, 3 families & 12 & 168 & 16.64\% & 11.39\% \\
Perception & Confidence only & 1 & 16 & -0.09\% & -0.33\% \\
Perception & Base, 3 families & 3 & 48 & 18.29\% & -0.86\% \\
Perception & Base+dense, 3 families & 6 & 120 & 19.72\% & 2.65\% \\
Perception & Own 4 views & 4 & 56 & 1.74\% & 1.15\% \\
Perception & 4 views, 2 families & 8 & 112 & 11.88\% & 0.99\% \\
Perception & 4 views, 3 families & 12 & 168 & 19.57\% & 2.11\% \\
\bottomrule
\end{tabularx}
\end{table}

\subsection{Repeated target splits}
\label{app:repeated}

The repeated splits use a response bank inferred again from the same three
models under the same four views, with per-question option probabilities
retained. Its prompt and inference runtime differ from the held-out runs, so
accuracies differ: Qwen3-VL answers 52.6\% of the HERBench reference questions
correctly, against 47.6\% in the held-out run, and 70.5\% of the Perception
Test reference questions, against 67.3\%. In the new HERBench bank, 9 of 208
video reads found no stored frame count; as in the held-out run, the count was
then estimated from duration and frame rate. The 30 further splits are fixed
random video assignments with the same pilot and test sizes as the reference
splits. All AURCs use the prefix-risk mean, including selection, and the
candidates, thresholds, and success settings are unchanged.
Tables~\ref{tab:repeated-perception} and~\ref{tab:repeated-herbench} give the
results by model.

On HERBench, the lowest split mean of the target-checked score is 10.54\%. The
check selects the target-pilot logistic score in 80 of the 90 evaluations and
the development HGB score in the other 10. On Perception Test, it keeps
InternVL's original score in 27 splits and never for LLaVA-OV2, whose minimum
gain is 22.24\%.

\begin{table}[H]
\centering
\caption{Mean relative AURC reductions (\%) by model over 30 further
Perception Test splits. Harm counts the model and split evaluations in which a
score ranks answers worse than the original score.}
\label{tab:repeated-perception}
\PaperTableSetup
\begin{tabularx}{\columnwidth}{@{\PaperTableEdge}Yrrrrr@{\PaperTableEdge}}
\toprule
Score & InternVL & Qwen & LLaVA & Mean & Harm \\
\midrule
All-view support & $-0.43$ & 9.96 & 26.00 & 11.84 & 17/90 \\
Base-view support & $-0.75$ & 11.51 & 26.84 & 12.53 & 22/90 \\
Target-checked score & 0.26 & 17.75 & 26.61 & 14.87 & 0/90 \\
\bottomrule
\end{tabularx}
\end{table}

\begin{table}[H]
\centering
\caption{Mean relative AURC reductions (\%) by model over 30 further HERBench
splits. Harm is counted as in Table~\ref{tab:repeated-perception}.}
\label{tab:repeated-herbench}
\PaperTableSetup
\begin{tabularx}{\columnwidth}{@{\PaperTableEdge}Yrrrrr@{\PaperTableEdge}}
\toprule
Score & InternVL & Qwen & LLaVA & Mean & Harm \\
\midrule
All-view support & 13.93 & 19.76 & 12.11 & 15.27 & 0/90 \\
Base-view support & 11.40 & 16.89 & 8.65 & 12.31 & 0/90 \\
Target-checked score & 10.84 & 19.33 & 10.83 & 13.66 & 0/90 \\
\bottomrule
\end{tabularx}
\end{table}

After these results were known, we added
all-view support to the check as a third candidate with a zero threshold,
dividing the 0.05 level among three candidates. This variant is exploratory and
not part of the prespecified rule. It selects support in 78 of the 90 HERBench
evaluations, never selects it for InternVL on Perception Test, and harms none of
the 180 evaluations. Its mean gain is 14.84\% on HERBench, 0.43 percentage points
below support alone, and 14.45\% on Perception Test, 0.42 points below the
two-candidate check.

We also tested majority voting over the three base answers, which, unlike the
method, changes the answers. A tied vote is resolved by the largest mean option
probability among tied options. On the 911 Perception Test questions of the
reference test split in the new bank, voting gives 70.80\% accuracy [67.34,
74.16], compared with 72.67\% [69.29, 75.86] for InternVL3.5, 70.47\% [67.10,
73.71] for Qwen3-VL, and 63.01\% [59.53, 66.37] for LLaVA-OV2. The paired
difference from InternVL3.5 is $-1.87$ percentage points [$-4.40$, 0.73]. All
intervals use the same 10,000 resamples of the 250 test videos.

\begin{table*}[t]
\centering
\caption{Scores evaluated with all VHD held out of training. Metrics are
averaged across the three development models, with 95\% source-video bootstrap
intervals. Gain is the relative AURC reduction from the original score;
negative values indicate worse ranking.}
\label{tab:vhd-heldout-scores}
\PaperTableSetup
\setlength{\tabcolsep}{3pt}
\begin{tabular}{lrr@{\hspace{12pt}}lrr}
\toprule
\multicolumn{3}{c}{All VHD} & \multicolumn{3}{c}{Visible} \\
\cmidrule(lr){1-3}\cmidrule(lr){4-6}
Score & AUROC & Gain (\%) & Score & AUROC & Gain (\%) \\
Original & $0.619\;[0.590, 0.646]$ & $0.00\;[0.00, 0.00]$ & Original & $0.636\;[0.592, 0.680]$ & $0.00\;[0.00, 0.00]$ \\
Base vote & $0.624\;[0.599, 0.647]$ & $-0.06\;[-2.62, 1.83]$ & Base vote & $0.639\;[0.600, 0.678]$ & $-1.28\;[-7.29, 2.15]$ \\
Base support & $0.624\;[0.594, 0.652]$ & $-3.31\;[-7.02, 0.11]$ & Base support & $0.615\;[0.567, 0.664]$ & $-16.00\;[-26.79, -7.11]$ \\
Logistic & $0.620\;[0.591, 0.648]$ & $-3.77\;[-7.46, -0.37]$ & Logistic & $0.603\;[0.554, 0.651]$ & $-17.59\;[-28.26, -8.71]$ \\
HGB & $0.633\;[0.601, 0.662]$ & $-2.14\;[-14.52, -0.09]$ & HGB & $0.603\;[0.556, 0.652]$ & $-19.17\;[-45.46, -10.02]$ \\
\midrule
\multicolumn{3}{c}{Counterintuitive} & \multicolumn{3}{c}{Hidden} \\
\cmidrule(lr){1-3}\cmidrule(lr){4-6}
Score & AUROC & Gain (\%) & Score & AUROC & Gain (\%) \\
Original & $0.384\;[0.315, 0.453]$ & $0.00\;[0.00, 0.00]$ & Original & $0.642\;[0.592, 0.693]$ & $0.00\;[0.00, 0.00]$ \\
Base vote & $0.311\;[0.236, 0.382]$ & $-6.36\;[-10.52, -3.03]$ & Base vote & $0.702\;[0.653, 0.753]$ & $10.88\;[4.85, 18.00]$ \\
Base support & $0.317\;[0.236, 0.396]$ & $-7.83\;[-14.04, -3.16]$ & Base support & $0.705\;[0.647, 0.764]$ & $8.37\;[-4.78, 19.99]$ \\
Logistic & $0.323\;[0.245, 0.397]$ & $-7.20\;[-13.28, -2.64]$ & Logistic & $0.715\;[0.656, 0.773]$ & $10.78\;[-2.44, 22.42]$ \\
HGB & $0.437\;[0.364, 0.511]$ & $-5.48\;[-9.16, -1.27]$ & HGB & $0.706\;[0.645, 0.768]$ & $12.48\;[-10.19, 22.37]$ \\
\bottomrule
\end{tabular}
\end{table*}

\subsection{Comparison with reconstructed PhaseGuard}
\label{app:phaseguard}

This exploratory comparison uses the Perception Test response bank and splits
above. Every scorer ranks the same answer, the option-probability maximum of
Qwen3-VL-8B-Instruct in the base view, so the results concern that answering
model alone. PhaseGuard reliability features are reconstructed from the
published description \cite{liu2026phaseguard}. On our four views, adjacency
and opposition refer to fixed view slots rather than literal phase offsets. We
compare both feature sets with the same fixed logistic regression and report the
complete target check alongside the reconstructed PhaseGuard selector and a
fixed random forest control (Table~\ref{tab:phaseguard-full}). The development
HGB candidate is the one trained on the original development data. Answer
voting and rescue are excluded. Both full scorers use 12 calls and 168 frame
exposures per question.

The two approaches are comparable, and none of the scorers harms any split. On
the reference split, the paired AURC difference between our features and
PhaseGuard features with the same logistic regression is $-0.00735$
[$-0.02007$, 0.00119]; negative values favor our features, and the interval
includes zero.

\begin{table}[H]
\centering
\caption{Mean relative AURC reductions and harmful outcomes for Qwen3-VL
answers under the four original views across 30 further Perception Test splits.
Harm means test AURC exceeds that of the original score. All PhaseGuard rows are
reconstructed controls.}
\label{tab:phaseguard-full}
\PaperTableSetup
\begin{tabularx}{\columnwidth}{@{\PaperTableEdge}Yrr@{\PaperTableEdge}}
\toprule
Score & Mean reduction & Harm \\
\midrule
Our features, logistic & 18.69\% & 0/30 \\
PhaseGuard features, same logistic & 18.07\% & 0/30 \\
Target-checked score & 17.75\% & 0/30 \\
PhaseGuard selected scorer & 18.15\% & 0/30 \\
PhaseGuard fixed random forest & 17.52\% & 0/30 \\
\bottomrule
\end{tabularx}
\end{table}

\subsection{VHD held out of training}
\label{app:vhd-holdout}

We reuse development logistic and HGB scores trained on NExT-GQA, STAR,
MVBench, and TempCompass, excluding all VHD variants. Base vote and base
support score the fixed answers, and base vote breaks ties with the original
score. Table~\ref{tab:vhd-heldout-scores} averages metrics across the three
development models, with intervals from 10,000 paired source-video resamples
that keep related variants and models together. On the 333 questions with
clearly visible evidence, the transferred scores also rank answers worse than
the original score, by 17.59\% for logistic regression and 19.17\% for HGB.
Shared-error percentiles are computed within each model's 178 counterintuitive
questions as 100 times the average rank divided by 178; higher percentiles
indicate greater trust.

Transfer also varies when the models stay the same. With the three development
checkpoints unchanged, we hold out each development dataset in turn and train
development logistic regression ($C=0.2$) and HGB on the remaining four, with
equal total weight per dataset (Table~\ref{tab:fixed-checkpoint}). Where HGB
uses an internal early-stopping split, that split keeps source videos together.
HGB gains range from 26.71\% on NExT-GQA to 1.49\% on MVBench and fall to
$-2.14$\% on VHD. Dataset content, task mix, and view budgets still differ
across the held-out datasets.

\begin{table}[H]
\centering
\caption{Development scores with fixed checkpoints, each development dataset
held out in turn. Values are mean relative AURC reductions across the three
development models, with 95\% source-video bootstrap intervals.}
\label{tab:fixed-checkpoint}
\PaperTableSetup
\begin{tabularx}{\columnwidth}{@{\PaperTableEdge}Yrr@{\PaperTableEdge}}
\toprule
Held-out dataset & Logistic & HGB \\
\midrule
NExT-GQA & 26.83\% [20.03, 34.04] & 26.71\% [11.88, 32.24] \\
STAR & 9.01\% [2.98, 15.41] & 11.32\% [2.23, 20.71] \\
MVBench & 13.76\% [10.60, 16.85] & 1.49\% [1.36, 12.24] \\
TempCompass & 12.24\% [6.24, 18.28] & 10.61\% [1.71, 17.64] \\
VHD & $-3.77$\% [$-7.46$, $-0.37$] & $-2.14$\% [$-14.52$, $-0.09$] \\
\bottomrule
\end{tabularx}
\end{table}

\end{document}